%% file: ms.tex
\documentclass{egpubl}
\usepackage{eg2021}
\ConferencePaper
\usepackage[T1]{fontenc}
\usepackage{dfadobe}
\usepackage{cite}
\biberVersion
\BibtexOrBiblatex
\electronicVersion
\PrintedOrElectronic
\ifpdf \usepackage[pdftex]{graphicx} \pdfcompresslevel=9
\else \usepackage[dvips]{graphicx} \fi
\EGpagenumber
\usepackage{egweblnk} 

\def\BibTeX{{\rm B\kern-.05em{\sc i\kern-.025em b}\kern-.08emT\kern-.1667em\lower.7ex\hbox{E}\kern-.125emX}}

\input{packages}
\title{Practical Face Reconstruction via Differentiable Ray Tracing}
\author
{\parbox{\textwidth}{\centering Abdallah Dib$^{\ast}$\;\;
                                Gaurav Bharaj$^{\dag\;\ddag}$\;\;
                                Junghyun Ahn$^{*}$\;\;
                                C\'edric Th\'ebault$^{*}$\;\;
                                Philippe-Henri Gosselin$^{*}$\;\;
                                Marco Romeo$^{\ddag}$\;\;
                                Louis Chevallier$^{*}$
        }
        \\
{\parbox{\textwidth}{\centering $^\ast$InterDigital R\&I\;\;\;\;\;
                                $^\dag$AI Foundation\;\;\;\;\;
                                $^\ddag$Technicolor Inc.
       }
}
\vspace{-20px}
}

\begin{document}
%
%
\teaser{
  \includegraphics[width=\textwidth]{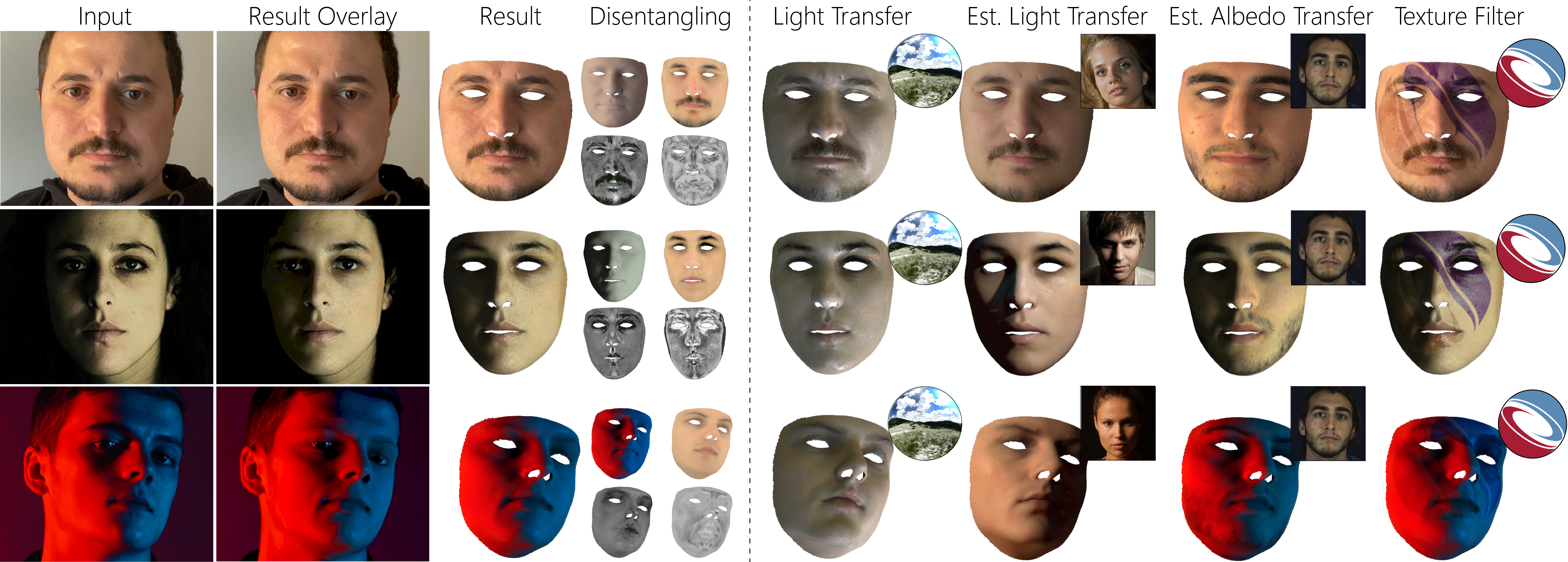}
  \caption{Our method takes as input an unconstrained monocular face image and estimates face attributes -- 3D pose, geometry, diffuse, specular, roughness and illumination (left). The estimation is self-shadow aware and handles varied illumination conditions. \eg{We show several resulting style transfer applications: albedos, illumination and textures transfers from and into face portrait images (right)}.}
  \label{fig:teaser}
  }
\maketitle
\input{0.abstract}
\vspace{-20px}
\input{1.introduction}
\vspace{-10px}
\input{2.relatedworks}
\vspace{-20px}
\input{3.problemformulation}
\vspace{-10px}
\input{4.optimization}
\vspace{-10px}
\input{5.results}
\vspace{-10px}
\input{6.ablation}
\vspace{-10px}
\input{6.comparisons}
\vspace{-10px}
\input{7.applications}
\vspace{-10px}
\input{8.conclusion}

\clearpage

\appendix
\input{9.appendix}
\bibliographystyle{eg-alpha-doi}  
\bibliography{ms}

\end{document}

%% file: packages.tex
\usepackage{amsfonts}
\usepackage{amsbsy}
\usepackage{amsmath}
\usepackage{microtype}
\usepackage[ruled,vlined]{algorithm2e}
\usepackage{xcolor}
\usepackage{makecell}
\usepackage{multirow}
\usepackage{adjustbox}
\usepackage{titlesec,titletoc}

\newcommand{\eg}[1]{\textcolor{black}{#1}}

\usepackage{bm}

\DeclareSymbolFont{Xlargesymbols}{OMX}{cmex}{m}{n}
\DeclareMathSymbol{\Xsum}{\mathop}{Xlargesymbols}{80}

%% file: 0.abstract.tex
\begin{abstract}
We present a differentiable ray-tracing based novel face reconstruction approach where scene attributes -- 3D geometry, reflectance (diffuse, specular and roughness), pose, camera parameters, and scene illumination -- are estimated from unconstrained monocular images. The proposed method models scene illumination via a novel, parameterized virtual light stage, which in-conjunction with differentiable ray-tracing, introduces a coarse-to-fine optimization formulation for face reconstruction. Our method can not only handle unconstrained illumination and self-shadows conditions, but also estimates diffuse and specular albedos. To estimate the face attributes consistently and with practical semantics, a two-stage optimization strategy systematically uses a subset of parametric attributes, where subsequent attribute estimations factor those previously estimated. For example, self-shadows estimated during the first stage, later prevent its baking into the personalized diffuse and specular albedos in the second stage. We show the efficacy of our approach in several real-world scenarios, where face attributes can be estimated even under extreme illumination conditions. Ablation studies, analyses and comparisons against several recent state-of-the-art methods show improved accuracy and versatility of our approach. With consistent face attributes reconstruction, our method leads to several style -- illumination, albedo, self-shadow -- edit and transfer applications, as discussed in the paper.
\end{abstract}

%% file: 1.introduction.tex
\section{Introduction}
\label{sec:intro}
\begin{figure}
\includegraphics[width=\linewidth]{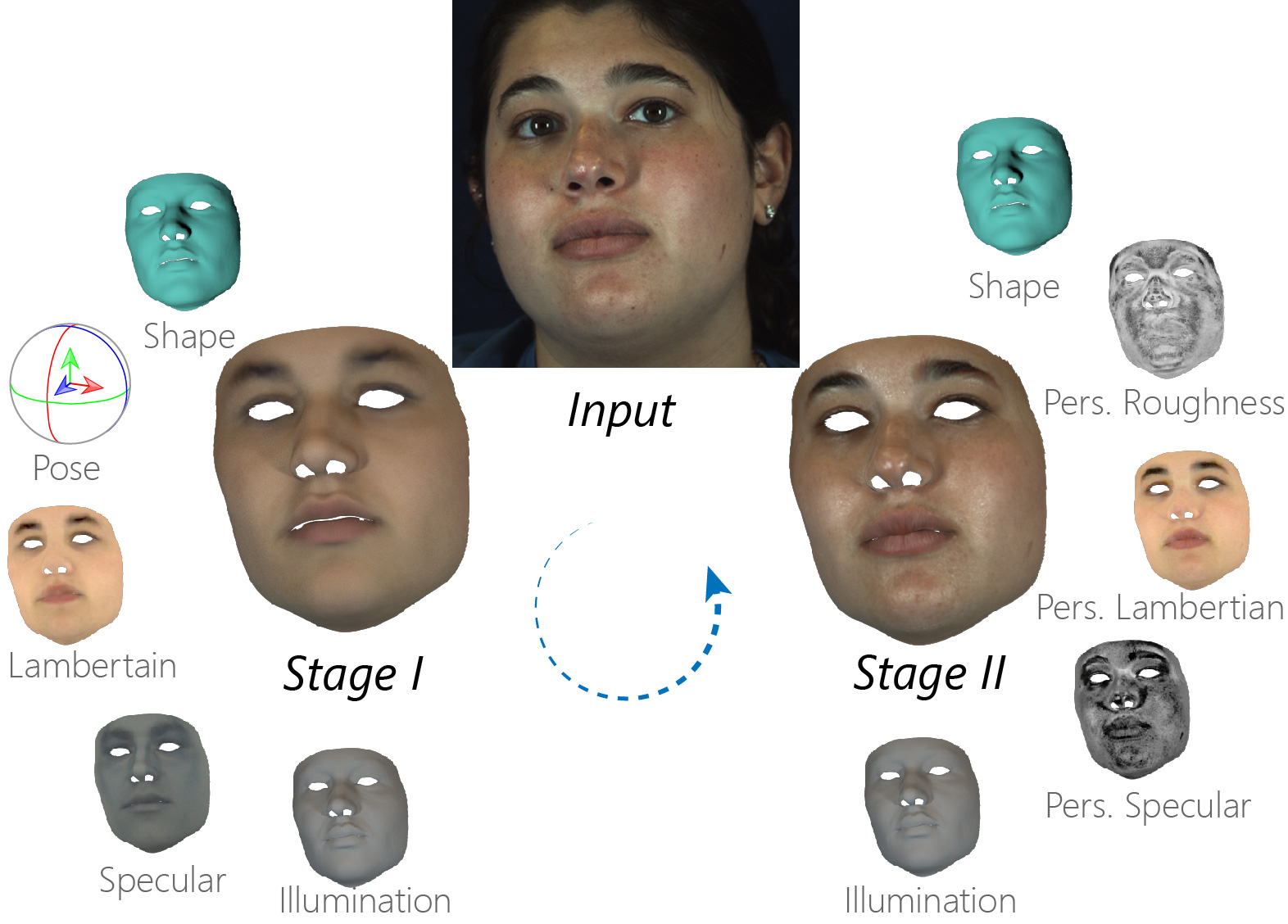}
  \caption{System Overview: Our method is divided into two stages. In Stage I, for an input image, geometry (pose, identity and expression), statistical diffuse $\mathsf{C}$ and specular $\mathsf{S}$ albedos and light stage illumination attributes are optimized. During this stage, the self-shadows are estimated as well. In Stage II, personalized diffuse $\mathsf{\hat{C}}$, specular $\mathcal{\hat{S}}$ and roughness $\mathcal{\hat{R}}$ attributes are estimated. Stage II takes into consideration attributes estimated in the previous stage.}
  \label{fig:overview}
  \vspace{-35px}
\end{figure}

Photorealistic \textit{avatarized} telecommunication, interactive AR/VR experiences and unobtrusive special effects for professional and consumer applications (e.g. \textit{selfie} filters) require accurate face reconstruction without specialized scene capture and subject/actor constraints. In several such \textit{in-the-wild} scenarios, users lack access to high quality and expensive camera and lighting hardware, or specialized personnel. For example, while interacting at-home through a monocular front facing camera, the user may encounter harsh self-shadows (for example, shadows cast by the nose or by the superciliary arch on the cheek), multicolored illumination or highly reflective skin conditions. 
Under varied conditions, consistent reconstruction of face attributes, while avoiding self-shadows biases, etc. is required. The method should work without manual intervention due to consumer constraints, while the reconstruction quality is on par with professional face motion capture systems.
\\
Monocular image-based face reconstruction with meaningful attributes estimation is hard due to its under-constrained nature. Given a face image, its pixel's final color values can be explained by several factors -- face shape, skin reflectance, camera position, or light color(s). This ambiguity makes it difficult to consistently estimate attributes. Unknown and unconstrained illumination conditions and consequent face self-shadows further add to the complexity. Our aim is to handle such scenarios using only monocular face images, while maintaining face reconstruction quality. This setup alleviates the need for specialized hardware and light requirements, that opens up avenues for movie production and VFX industry scenarios.
\\
Face reconstruction methods ~\cite{zollhoefer2018facestar,tran2018nonlinear,tewari2019fml,sanyal2019learning} estimate geometry based on parametric face models -- 3D morphable model (3DMM)~\cite{Egger20Years}. 
Such methods assume Lambertian skin reflectance~\cite{angel2012interactive} with distant light illumination, where the incoming radiance is a function of direction. 
Under this assumption, \textit{spherical harmonics} \cite{ramamoorthi2001relationship} have been widely used to model scene illumination. These methods do not model self-shadows. The projected face shape's \textit{geometry-patch} corresponding to color saturated (due to shadows, albedos, illumination) pixel patches can lead to unnatural geometric deformations and inconsistent attribute estimation. More recently, \cite{shu2017neural,yamaguchi2018high,smith2020morphable,lattas2020avatarme} introduce specular reflectance modeling based data-driven priors, however, they do not explicitly handle self-shadows. While, more complete controlled face reconstruction methods \cite{debevec2000acquiring,ghosh2009estimating,gotardo2018practical} exist, such methods are not applicable for at-home consumer, unobstructed and live performance capture scenarios, due to extensive hardware requirements, and set pre-conditions.
\\
Our objective is 3D face reconstruction with explicit separation of face attributes -- skin reflectance (diffuse, specular and roughness), 3D geometry (identity and expression), pose and illumination -- from input images. To this end, we use statistical 3DMM to model base face geometry, diffuse and specular albedos priors, along with Cook-Torrance bidirectional scattering distribution function (BRDF) \cite{Schlick94aninexpensive} to model skin reflectance. Each vertex on the geometry is characterized by diffuse, specular and roughness parameters; illumination is modeled via a novel virtual light stage with parameterized lights. We also obtain personalized albedos, that refine the statistical 3DMM-based initial estimates. Modeling parameters are used to synthesize an image using differentiable ray tracing, that also obtains self-shadows. Input and synthesized images are used to minimize a photo-consistency loss in two stages, where each stage minimizes a subset of the parameters. We note that although more accurate and complete reflectance modeling approaches~\cite{weyrich2006analysis} exist, given the quality and nature of input images, the Cook-Torrance reflectance model suffices for our reconstruction needs. 
\\
Face attribute reconstruction from monocular images is highly nonlinear, our experiments show that naively optimizing all the parameters jointly can lead to poor results. Optimized jointly, specular albedo may get \textit{baked} into diffuse albedo, shadows, etc. Thus, a better strategy for attributes reconstruction is required. We introduce a two stage optimization (Figure~\ref{fig:overview}), where in first stage, similar to~\cite{Garrido:2016, smith2020morphable} we optimize the pose, illumination, geometry, diffuse and specular albedos, statistically regularized by the 3DMM, while specular roughness remains fixed. Due to ray tracing, the interplay between estimated geometry and illumination helps extract self-shadows. At this stage, person specific (from input image) face attributes such as facial hair, moles, etc. are not estimated. In second stage, we extract unconstrained diffuse, specular and roughness that captures person specific facial details not modeled via statistical diffuse or specular albedos. This staged optimization strategy adds structure and makes the under-constrained optimization problem tractable, leading to superior reconstruction vs. the naive approach. To summarize, the main contributions of our work include:
\begin{itemize}
\item A novel virtual light stage formulation, which in-conjunction with differentiable ray tracing, obtains more accurate scene illumination and reflectance, implicitly modeling self-shadows. The virtual light stage models, the switch from point to directional area lights and vice-versa, Sec.~\ref{sec:facemodelingformulation}.
\item Face reflectance -- diffuse, specular and roughness reconstruction that is scene illumination and self-shadows aware.
\item A robust optimization strategy that extracts semantically meaningful personalized face attributes, from unconstrained images, Sec.~\ref{sec:optimization}. 
\end{itemize}
To demonstrate the efficacy of our approach we provide several results (Sec~\ref{sec:results}), ablation (Sec~\ref{ssec:ablation}) and extensive comparisons (Sec~\ref{sec:comparisons}) against state-of-the-art methods, where geometric, diffuse and specular albedo estimates are compared. We also compare the proposed light-stage formulation against high-order spherical-harmonic light modeling. Since our method provides fine control over the face attributes, it leads to several style edition and transfer applications (Sec~\ref{sec:applications}) such as face portrait relighting, illumination transfer, specular reflections and self-shadow editing, etc. Scenarios such as changing face pose with accurate resultant self-shadows, or changing illumination, or addition of face \textit{texture} filters, while maintaining original specular albedo (Figure~\ref{fig:teaser}), are possible. Finally, in Sec~\ref{sec:conclusion} we conclude with limitations and future works.

%% file: 2.relatedworks.tex
\section{Related Works}
\label{sec:relatedworks}
Face reconstruction from single, multi-camera images, videos or time-of-flight depth data, is a classic computer vision problem, where the goal is accurate geometry and reflectance reconstruction. With rapid progress in mobile camera technologies, \textit{selfie}-photography, social media, and telecommunication applications, \textit{single camera} face reconstruction approaches has gained special attention. Camera depth ambiguity, capture conditions, non-convexity of face shapes, reflectance properties of human skin, shadows, and illumination conditions make monocular face reconstruction extremely challenging. Several methods have been proposed, that solve for a subset of the face attributes -- 3D geometry (neural shape and expressions), pose, diffuse, specular, roughness and illumination (including self-shadows).
\vspace{4px}
\\
\textit{Geometry and Reflectance Modeling.} \cite{beeler2011high} presents multi-view camera and controlled illumination based photogrammetric method that produces high-quality (includes mesoscopic face details) temporally stable face geometries. \cite{wu2011shading, valgaerts2012lightweight} propose a stereo-image methods for face reconstruction and shape-from-shading based geometry refinement. \eg{\cite{ghoshPractical} captures high-fidelity and multilayered face reflectance using (single camera) multiple images without other external hardware. \cite{grahamNearInstant} captures high quality face geometry and reflectance (diffuse, specular) via a multiview camera setup. More recently, \cite{riviere2020SingleshotHF} present a lightweight low-cost rig for high-quality acquisition of facial geometry and appearance with fine-scale pore details.}
\\
Photogrammetric and external hardware based approaches provide extremely accurate results, but add constraints on the capture scenarios: multi-view cameras, extensive illumination setups, or lighting conditions (e.g. no self-shadows) for optimal capture. With such approaches if a single camera is used, the reconstruction formulation has infinite deformation degrees-of-freedom, making the problem infeasible. Thus, such methods are not applicable for \textit{in-the-wild} monocular images. Most of these methods do not model specular reflectance and assume a diffuse skin reflectance model.
\\
In-order to use unconstrained monocular images, statistical priors have been introduced \cite{zollhoefer2018facestar}. Such priors add structure to the reconstruction formulation. 3D Morphable Models (3DMMs) \cite{blanz1999morphable,li2017learning,Egger20Years} use facial scanning hardware to capture ground-truth geometry and (diffuse) reflectance. Later, dimensionality reduction method such as principles component analysis (PCA) is used to create linear parametric models. \cite{ Garrido2013,suwajanakorn2014total,Garrido:2016} introduce optimization formulation for geometry (and diffuse) reflectance reconstruction, where 3DMM based priors act as optimization regularizer. They estimate camera parameters and minimize photo-consistency losses based on input images. Such methods also use sparse face image features such as landmarks~\cite{saragih2011deformable}, that \textit{regularize} the optimization against local minima. In-order to separate neutral face shape from expression, FACS~\cite{ekman1997face} based blendshapes PCA models are used. These methods work well for controlled scene conditions, and often do no generalize well for \textit{in-the-wild} images scenarios. Where they can bake shadows, specularity into diffuse albedo and vice-versa. 
\\
\cite{LiChenIntrinsics} extracts diffuse and specular albedos from a single image using Spherical Harmonics (SH) illumination, however, they do not model explicit self-shadows. \cite{tewari17MoFA, tewari2019fml} use self-supervised autoencoders and inverse-rendering architectures to \textit{infer} 3DMM's linearized semantic attributes. Nonlinear face geometry models such as mesh autoencoders~\cite{ranjan2018generating} and CNN encoder~\cite{tran2018nonlinear} have also been proposed. Using high quality face datasets and novel deep learning algorithms, \cite{saito2017photorealistic,laine2017production,bagautdinov2018modeling} show vast improvements in geometry reconstruction. \cite{huynh2018mesoscopic} shows further improvements by inferring mesoscopic facial attributes given monocular images, 
an attribute we do not model in our reconstruction approach.
\\
True human skin reflectance capture and reconstruction is a hard problem and several BRDF-based \cite{nicodemus1992geometrical} formulations have been proposed. \cite{takiwaki1998measurement,debevec2000acquiring,weyrich2006analysis,alexander2010digital,ghosh2011multiview} propose extensive measurement systems, structured light setups and data-driven methods. While such approaches lead to highly accurate skin (diffuse and specular) reflectance modeling, they require controlled capture conditions and extensive calibration. Our aim, instead, is to robustly extract face attributes from unconstrained images, 
where a highly accurate skin reflectance models may not be applicable due to the low quality of input images.
\cite{gotardo2018practical} provides a more practical approach to model skin reflectance and ambient occlusions-based shading. Although their setup is less extensive than other approaches, it still requires a controlled multi-view and multi-light illumination setup for reflectance modeling.
\\
Most face reconstruction approaches rely on a lightweight parametric skin reflectance model using linear Lambertian models, where it is assumed that skin \textit{does not} have specular attributes. This simplification has shown great success for face reconstruction~\cite{Garrido:2016,tewari17MoFA,sengupta2018sfsnet}. Recently, \cite{yamaguchi2018high,smith2020morphable,lattas2020avatarme} add specular (without roughness) reflectance modeling from unconstrained images, as a result the extracted face models have better attribute disentangling. These methods are more robust against strong self-shadows and specular reflections in input images. However, as discussed in Section~\ref{sec:comparisons}, they do not \textit{fully} estimate face attributes under several illumination scenarios and bake these attributes in diffuse and specular albedos. While \cite{yamaguchi2018high, lattas2020avatarme} infer geometry and reflectance, but not the illumination. Self-shadows baked into the albedos can be observed, whereas we model self-shadows implicitly.
\vspace{2px}
\\
\textit{Illumination modeling.} Scene illumination can be modeled via light probes~\cite{reinhard2010high, legendre2016practical}, environment maps~\cite{hakura2001parameterized}, \eg{sparse mixture of spherical gaussians~\cite{OM3D2014}}, and illumination model relying on Spherical Harmonics~\cite{ramamoorthi2001relationship} that assume Lambertian reflectance. While illumination capture requires specialized hardware, having a linear illumination model limits attributes separation such as self-shadows. 
Most approaches assume that illumination is mostly uniform resulting in self-shadow being baked into albedo attribute.
One way to approach this limitation is to \textit{mask} shadowed patched via occlusion maps, and use GANs~\cite{nagano2018pagan} to fill-in the albedos. We approach this problem from a different perspective, similar to initial experiments by \cite{dib2019face} a novel parameterized virtual area light stage is introduced that simulates real world illumination conditions. This illumination model is used together with ray tracing, that implicitly models self-shadow attributes. Consequently, it reconstructs geometric patch's reflectance separating incurred shadows (Sec \ref{ssec:IlluminationModeling}). To the best of our knowledge, the proposed method is the first to estimate reflectance (diffuse, specular, roughness), illumination, and self-shadows robustly from monocular images. 
\vspace{2px}
\\
\textit{Applications.} High-quality face reconstruction leads to several use cases for consumer and movie production scenarios. While quality face tracking has several advantages, such as reenactment, realistic virtual avatars~\cite{suwajanakorn2017synthesizing,kim2018deep}, attributes separation  opens up new possibilities. Photoshop-like applications for face portrait touch-up have been proposed. For example, \cite{shih2014style} shows how \textit{style} from one image can be transferred to another employing image-based methods for style transfer. \cite{shu2017portrait} proposes a method for illumination transfer from source to target images, while \cite{sun2019single} describes a method for portrait relighting. 
More recently, \cite{zhang2020portrait} proposes a method for foreign shadow removal from images. Since our method can separates several face attributes, it makes many such applications feasible, as discussed in the paper.

%% file: 3.problemformulation.tex
\vspace{8px}
\section{Face Modeling Formulation}
\label{sec:facemodelingformulation}
\textbf{Overview.} We propose a practical formulation to model and reconstruct face attributes. Sec~\ref{ssec:GeometryModeling} describes geometry modeling, and Sec~\ref{ssec:ReflectanceModeling} describes parameterized reflectance model for diffuse and specular albedo modeling using statistical priors and Cook-Torrance model for personalization. Sec~\ref{ssec:IlluminationModeling} introduces our novel parameterized virtual light stage for scene illumination modeling with differentiable ray-tracing. These parametric attributes are then formulated in Sec~\ref{sec:optimization} into an optimization, and solved with a new two-stage optimization strategy (Fig~\ref{fig:overview}). 
\vspace{-10px}
\subsection{Geometry Modeling}
\label{ssec:GeometryModeling}
Similar to~\cite{Garrido:2016}, geometry is modeled via 3DMM and photo-consistency loss. This loss is regularized via a sparse set of face landmarks, where we employ state-of-the-art 2D landmarks estimation~\cite{bulat2017far}. This sparse landmark loss (Section~\ref{sec:optimization}), helps regularize against local minima where photo-consistency loss is under-constrained, especially under low light, heavy specular or self-shadow conditions. 
We use \cite{blanz1999morphable, gerig2018morphable}'s statistical face model, where identity is given by $\mathsf{e} = \mathsf{a}_s + \mathsf\Sigma_s \mathsf{\alpha}$. $\mathsf{e}$ a vector of face geometry vertices with $|\mathsf{e}| = N$. The identity space is spanned by $\mathsf\Sigma_s \in \mathbb{R}^{3N \times K_s}$ composed of $K_s = 80$ principal components of the identity shape-space. $\mathsf{\alpha} \in \mathbb{R}^{K_s}$ describes weights for each coefficient of the 3DMM and $\mathsf{a}_s\in\mathbb{R}^{3N}$ is the average face mesh. We model face expressions over the neutral identity by $\mathsf{e}$ via linearized blendshapes $\mathsf{v} = \mathsf{e} + \mathsf\Sigma_e \mathsf{\delta}$, where $\mathsf{v}$ is the final vertex position displaced from $\mathsf{e}$ by weight vector $\mathsf{\delta} \in \mathbb{R}^{ K_e} $ and $\mathsf\Sigma_e \in \mathbb{R}^{3N \times K_e} $ containing $K_e = 75$ principal components of the expression space.
\vspace{4px}
\\
\textbf{Camera model.} We use a pinhole camera model with rotation $\mathsf{R} \in \mathsf{SO}(3)$  and translation $\mathsf{T} \in \mathbb{R}^3$. We assume the camera is always centered at the origin and $ \Gamma(\mathsf{v}_i) = \mathsf{R}^{-1}(\mathsf{v}_i -\mathsf{T})$ is the transformation that maps a vertex $\mathsf{v}_i\in\mathbb{R}^3$ to the camera coordinate frame. $\Pi$ is the perspective camera matrix that maps a 3D vertex to a 2D pixel.
\begin{figure}
\includegraphics[width=\linewidth]{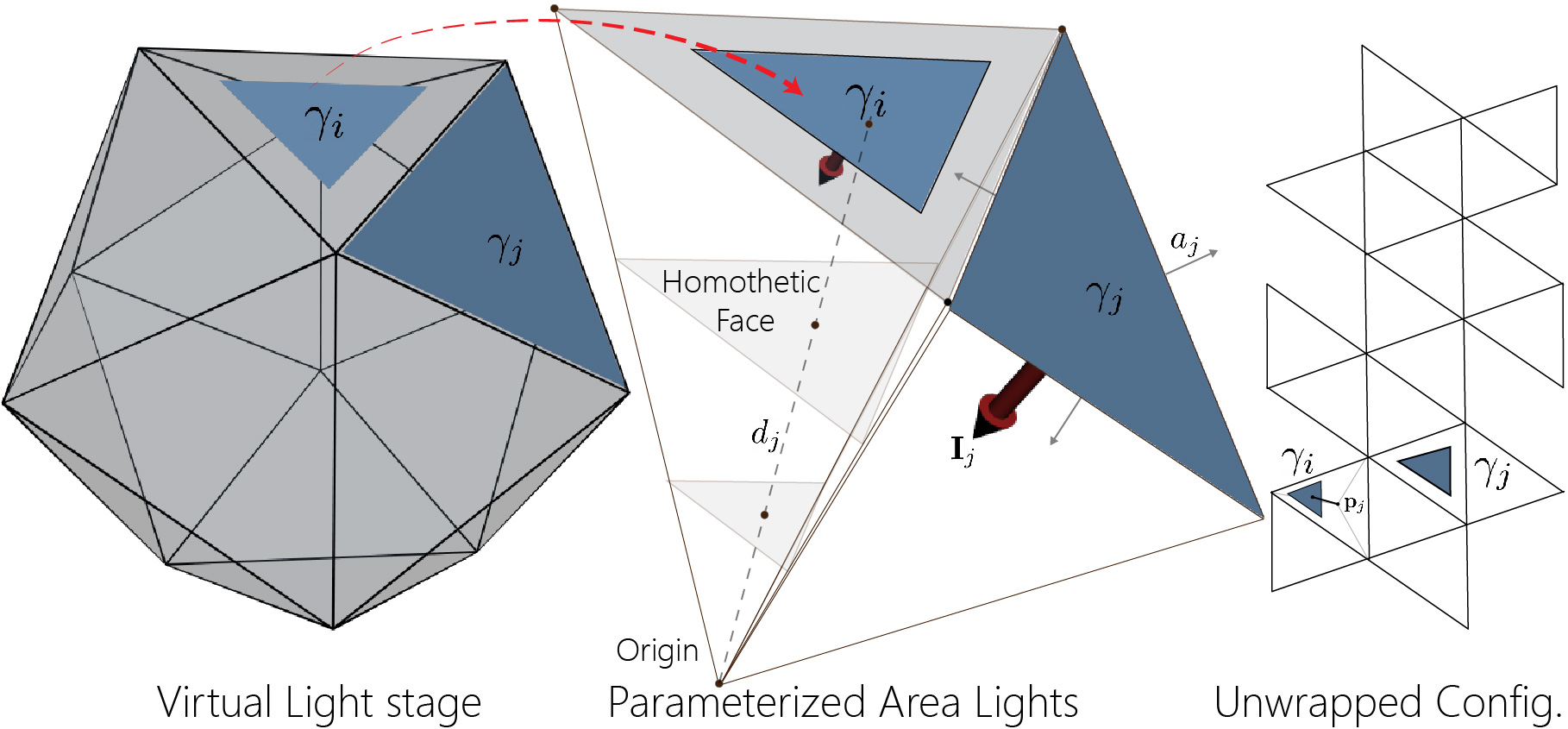}
\caption{Left: Our virtual light stage has an icosahedronic geometric construction. Middle: From each of the twenty faces of the icosahedron, we create: parameterized area lights $\gamma_j$ with intensity $\mathsf{i}_j$, surface area $\mathsf{a}_j$, position $\mathsf{d}_j$ and distance to the origin $\mathsf{d}_j$. Right: Unwrapped representation of the icosahedron.}
\label{fig:lightstage}
\vspace{-20px}
\end{figure}
\vspace{-10px}
\subsection{Reflectance Modeling}
\label{ssec:ReflectanceModeling}
We use Cook-Torrance BRDF~\cite{cook1982reflectance,walter2007Microfacet} to model face skin reflectance, that defines for each geometry vertex $\mathsf{v}_i$: a diffuse (color) $\mathsf{c}_i \in \mathbb{R}^3$, specular $\mathsf{s}_i\in \mathbb{R}^3$ and roughness $r_i \in \mathbb{R}$ albedos. The BRDF model that defines how the incoming light is reflected on the surface geometry is given by: 
\begin{align}
 \label{eq:brdf}
 f_r(\mathsf{s}_i,r_i, \mathsf{c}_i, \mathsf{n}_i, \mathsf{l}, \mathsf{o}) &= f_{d}(\mathsf{c}_i) + f_{s}(\mathsf{s}_i, r_i, \mathsf{n}_i, \mathsf{l}, \mathsf{o})
 \end{align}
 $f_{d}$ is the material term for diffused light in all directions. $f_{s}$ is the specular term for light reflected for a given viewing direction. In contrast to Lambertian BRDF model, the Cook-Torrance BRDF allows us to model specular highlights on the skin's surface. 
 $\mathsf{n}_i\in \mathbb{R}^3$ is the normal at vertex $\mathsf{v}_i$ and $\mathsf{l}\in\mathbb{R}^3$ is the incident area light direction (Section \ref{ssec:IlluminationModeling}). $\mathsf{o}\in\mathbb{R}^3$ the view direction pointing to the origin of the pinhole camera. For a quick refresher on $f_{d}$ and $f_{s}$ reflectance modeling, we refer interested reader to the supplementary material (section A).
\\
The statistical diffuse albedo $\mathsf{c}\in\mathbb{R}^{3N}$ is derived from 3DMM as $\mathsf{c} = \mathsf{a}_r + \mathsf\Sigma_r \mathsf\beta$, where $\mathsf\Sigma_r\in\mathbb{R}^{3N \times K_r}$ defines the PCA diffuse reflectance with $K_r=80$ and $\mathsf\beta\in\mathbb{R}^{K_r}$ the coefficients. $\mathsf{a}_r$ is the average skin diffuse reflectance. Similarly, we employ the statistical specular prior introduced by \cite{smith2020morphable} to model the specular reflectance: $\mathsf{s} = \mathsf{a}_b + \mathsf\Sigma_b \mathsf\gamma$ where $\mathsf\Sigma_b\in\mathbb{R}^{3N \times K_b} $ defines the PCA specular reflectance with $K_b=80$ and $\mathsf\gamma\in\mathbb{R}^{K_b}$ as the coefficients.  $\mathsf{a}_b$ is the average specular reflectance. Note that, \cite{smith2020morphable} recommends using $\mathsf{\gamma} = \mathsf{\beta}$, however, we use separate parameterization with regularization that leads to similar results with more flexibility. 
\\
In unwrapped (UV) image texture space, $\mathcal{C} \in \mathbb{R}^{M \times M \times 3}$ and $\mathcal{S} \in \mathbb{R}^{M \times M\times3}$ are the statistical diffuse and specular albedos, respectively. $\mathcal{\hat R} \in \mathbb{{R}}^{M \times M}$ defines roughness (no given statistical prior), with $M \times M$ texture resolution. For each projected vertex onto the texture, $\mathcal{C}, \mathcal{S}$ and $\mathcal{\hat{R}}$ describes the interpolated $(r, g, b)$ color, specularity and roughness factors for vertex $\mathsf{v}_i$, where, statistical diffuse albedo $\mathsf{c}_i = \mathcal{C}(\mathfrak{u}_i, \mathfrak{v}_i)$, statistical specular albedo $\mathsf{s}_i = \mathcal{S}(\mathfrak{u}_i,\mathfrak{v}_i)$, roughness $r_i = \mathcal{\hat{R}}(\mathfrak{u}_i, \mathfrak{v}_i)$. $\{\mathfrak{u}_i, \mathfrak{v}_i\} \in [0, 1]$ is projection of vertex $\mathsf{v}_i$ onto $UV$ space.
\vspace{4px}
\\
\textbf{Image-based Personalized Albedo.} In Stage I (Section \ref{sec:optimization}), statistical diffuse $\mathcal{C}$ and specular $\mathcal{S}$ albedos are constrained by 3DMM. In Stage II, we personalize albedos using the input image to capture person specific details -- facial hair, moles, coloration, and oiliness. Thus, Stage II refines the initially estimated (Stage I) albedo for unconstrained diffuse $\mathcal{\hat{C}}$, specular $\mathcal{\hat{S}}$, and additionally roughness $\mathcal{\hat{R}}$.
\vspace{-10px}
\subsection{Illumination Modeling}
\label{ssec:IlluminationModeling}
Introduced by~\cite{ramamoorthi2001relationship}, spherical harmonics (SH), is a method for illumination modeling (assumes light at infinity) with Lambertian reflectance. \cite{debevec2000acquiring} introduces a method to capture scene light, that can be used as an environment maps for image-based lighting. \cite{gortler1996lumigraph} introduced Lumigraph, to model a complex 4D plenoptic function that describes the flow of light at all positions in all directions for a given scene. Some of these methods require physical apparatus, some are parametrically complex, while others introduce material modeling limitations. In our initial experiments, we formulated illumination modeling using both higher-order SH and environment maps. However, these methods result in sub-optimal self-shadows modeling, and attribute disentangling (see Section~\ref{ssec:ablation}).
\\
For our problem, we need a lightweight yet flexible, parametric scene illumination approach that can not only approximate incoming light, but also model bright, dim, non-uniform, multi-color illumination over non-convex face geometry. Moreover, unlike SH and environment maps, 
we want to model semantically meaningful light configurations such as point, area, and directional. Thus, we introduce the \textit{virtual light stage} illumination model. For physical face geometry capture, structured light approaches ~\cite{ghosh2009estimating} exist, such methods build physical rigs, known as light stages, with programmable lights and cameras. Inspired from light rigs, we form our virtual light stage that loosely simulates these physical structures to model scene illumination.
\\
To model incoming light on face geometry, we explore various geometric configurations such as a tetrahedron, octahedron, icosahedron and spherical -- convex 3D manifolds. Such configurations' triangles can be thought of as area lights, directed towards the manifold's origin. In our experiments, we observe that these light stage configurations practically satisfy the requirements for inctoming light needed for face modeling. During our nascent explorations, we tried very simple structures such as a tetrahedron with four area lights, and more complex geometries like discrete sphere with eighty area lights. Along the various geometric structures, \textit{icosahedron} provides optimal complexity for illumination modeling. See Section~\ref{ssec:ablation} for comparisons and Supplementary material for various configurations and resultant face reconstructions.
\vspace{4px}
\\
\textbf{Virtual Light Stage.} A virtual light stage with area lights $\gamma_j$, $j \in \{1, ..., 20\}$, an icosahedron is shown in Fig~\ref{fig:lightstage}. The shape, size and position of the area lights are derived from the face triangles of the icosahedron. Each area light, modeled independently, has the following parameters: distance $\mathsf{d}_j \in\mathbb{R}$ from the face geometry (at the origin), relative surface area $\mathsf{a}_j \in\mathbb{R}$, local position $\mathsf{p}_j \in\mathbb{R}^2$ of the light center in barycentric coordinates within the face triangle, and perceived intensity $\mathsf{i}_j \in\mathbb{R}^3$. We define $\mathsf\gamma_j = \{ \mathsf{d}_j, \mathsf{a}_j, \mathsf{p}_j, \mathsf{i}_j \}$ as the set of parameters for an area light. Each light can be \textit{switched-off} by setting the perceived intensity parameter $\mathsf{i}_j$ to zero. The physical intensity $\mathsf{I}_j \in\mathbb{R}^3$ used for illumination is given by:
\vspace{-8px}
\begin{align}
\label{eq:lightIntensity}
\mathsf{I}_j = \frac{\mathsf{d}_j^2}{\mathsf{a}_j} \mathsf{i}_j
\end{align}
Here, the surface area $\mathsf{a}_j$ of the light is relative to the face triangle's area. 
$\mathsf{a}_j$ is bound between 0 -- corresponding to a point light, and 1 -- maximum surface area of the face triangle. This parameter set has been chosen to better decouple the light parameters. With the standard illumination equation, the light influx reaching an object depends on the physical intensity, distance and size of the light. But, our formulation decouples these parameters and makes it possible to operate only on a single light parameter without effecting other parameters. These variables are orthogonal, and ease the optimization. Without this orthogonal representation, if the effect of a light is too strong, the optimization would have several degrees-of-freedom to change intensity, such as position, size of the light, etc., while, in our formulation, only parameter $i_j$ is needed to modify intensity.
\\
During the initialization, an area light is positioned at center of each triangle of the light stage icosahedron. Each light $\gamma_j$ can move according to its distance $\mathsf{d}_j$ from the geometry center -- its size remaining proportional to $\mathsf{d}_j$. $\mathsf{a}_j$ and $\mathsf{p}_j$ are used to control position and size of each light $\gamma_j$ within the surface defined by the homothetic face -- the icosahedron face scaled by $\mathsf{d}_j$. Thus, the area light remains parallel to original icosahedron's face. A soft box constraint ensures the area lights stay within these homothetic faces (see Section \ref{sec:optimization}). The position and size of the area light control incident light beams, and thus determine the position and the appearance of self-shadows -- soft or hard, and specular reflections. 
When the lights share identical parameters, they are uniformly distributed over 3D angular space; in this case, the model can approximate uniform illumination. The surface of an area light can also become small enough to approximate point light sources.
\vspace{4px}
\\
\textbf{Shadows approximation.} In Section~\ref{sec:optimization}, we introduce our optimization formulation that relies on differentiable ray tracing for image synthesis. By varying the number of ray-\textit{bounces} against scene geometries and subsequent indirect illumination, self-shadows can be modeled. That is, \textit{gradient} of shading for a geometric face is dependent on the ray bounces that contribute to incoming light on a face. In our formulation, since we have no information on scene geometry (other than the human face), we do not model indirect illumination due to lack of geometry to bounce-off from. We avoid self-geometry bounces, as in our experiments, it did not lead to substantial gains in accuracy. By using area lights that can be turned \textit{on} or \textit{off}, and by controlling their intensity, position and surface area, we are capable of modeling several illumination and self-shadow scenarios.

%% file: 4.optimization.tex
\section{Optimization}
\label{sec:optimization}
\begin{figure*}
    \centering
    \includegraphics[width=\linewidth]{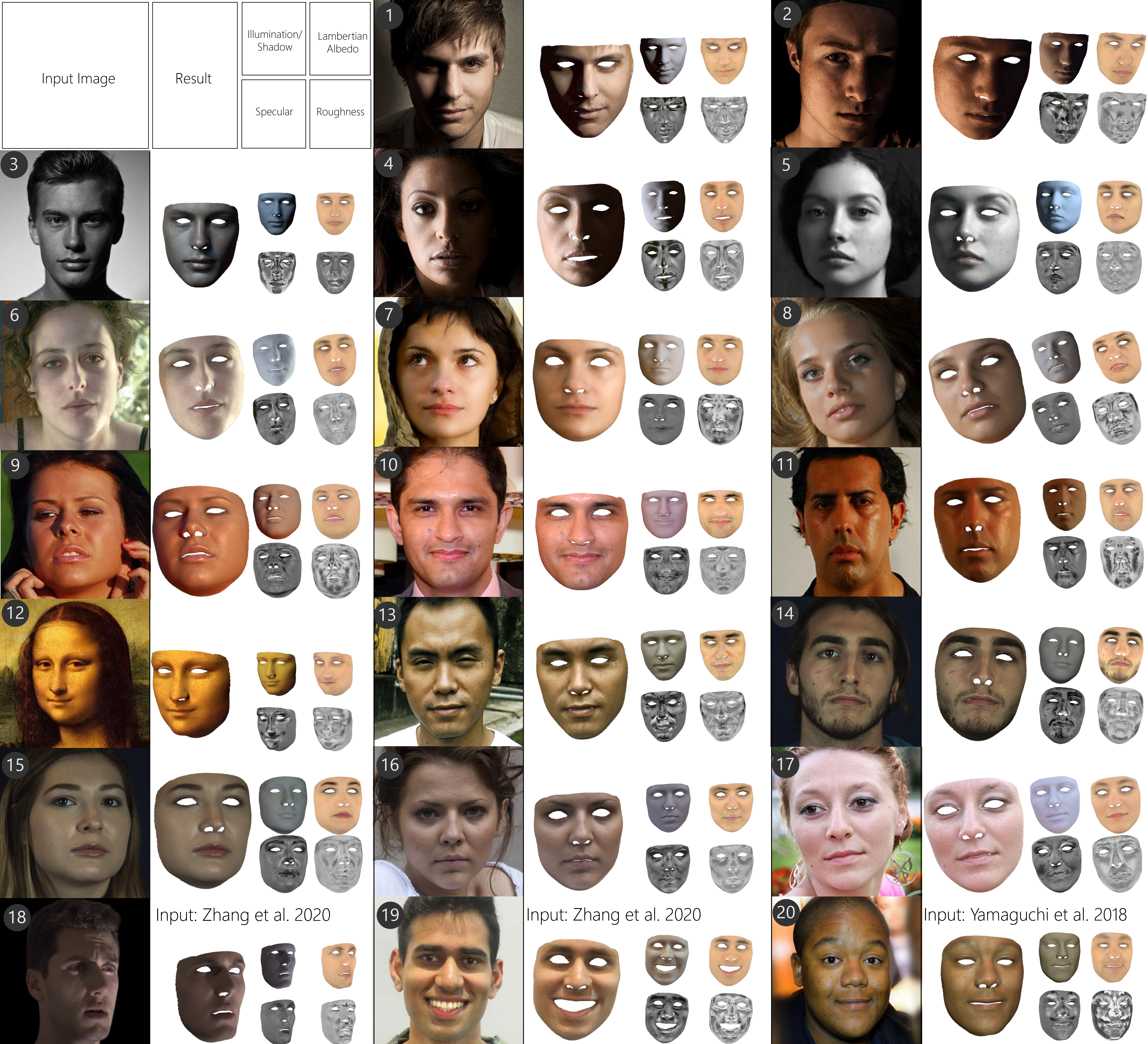}
  \caption{For each image we show the final optimization result with the estimated parameters: illumination (with estimated self-shadows), diffuse, specular albedo and roughness.}
  \label{fig:results}
  \vspace{-20px}
\end{figure*}
Our goal is robust face reconstruction via geometry (pose, identity and expression), reflectance (diffuse, specular, roughness) and illumination estimation. With unconstrained illumination the optimization can become under-constrained,  we therefore resort to a carefully designed two staged optimization strategy. In each stage, Figure \ref{fig:overview}, we select a subset of the face attributes. Our \textit{analysis-by-synthesis} approach consists in synthesizing an image using parameters $\chi = \{\mathsf\omega, \mathsf\alpha, \mathsf\delta, \mathsf\beta, \mathsf\gamma, \mathsf{R}, \mathsf{T}\}$ (where $\mathsf\omega = \{ \mathsf{d}, \mathsf{a}, \mathsf{p}, \mathsf{i} \}$ are the light stage parameters) using differentiable ray tracing~\cite{Li:2018:DMC}. This minimizes a photo-consistency loss between synthesized $\mathcal{I}^S$ and real $\mathcal{I}^R$ images on per pixel basis:
\begin{equation}
 \label{eq:energy}
\mathsf{E}_{ph}(\chi) =\Xsum_{i\in\mathcal{I}} |\mathsf{p}_i^S(\chi) - \mathsf{p}_i^R|
\end{equation}
Here, $\mathsf{p}_i^S, \mathsf{p}_i^R \in\mathbb{R}^3$ are ray traced and real image pixel colors, respectively. Rendered pixel colors are given by $\mathsf{p}_i^S=\mathcal{F}(\mathsf\omega, \mathsf\alpha, \mathsf\delta, \mathsf\beta, \gamma, \mathsf{R}, \mathsf{T})$, where $\mathcal{F}$ is the Monte Carlo estimator of the rendering equation \cite{Kajiya86Rendering}. We also define a sparse landmark loss that measures the distance between the projection of $L = 68$ facial landmarks and their corresponding pixel projections $\mathsf{z}_l$ on input image:
\vspace{-5px}
\begin{equation}
 \label{eq:landmark}
\mathsf{E}_{land}(\chi) =\Xsum_{l = 1}^{L} || \Pi \circ \Gamma(\mathsf{v}_i^l) - \mathsf{z}_l||_2^2
\end{equation}
The sparse landmark loss regularizes the optimization against local minima occuring when photo-consistency loss is ambiguous.
\vspace{4px}
\\
\textbf{Optimization strategy.} We introduce a two-stage optimization strategy, where Stage I uses statistically regularized albedo priors and Stage II optimizes unconstrained albedos:
\\
\textbf{Stage I.} We optimize camera parameters $\Gamma$ and blendshape coefficients using the landmark loss (Eq~\ref{eq:landmark}). After this pose and expression initialization, we introduce the optimization for statistical albedos ($\mathsf\beta$ and $\mathsf\gamma$), face geometry and expression ($\mathsf\alpha, \mathsf\delta$), illumination ($\mathsf\omega$), and camera ($\mathsf{R}, \mathsf{T}$), while other parameters -- specularity $\mathcal{\hat{S}}$, roughness $\mathcal{\hat{R}}$ and diffuse albedo $\mathcal{\hat{C}}$ -- remain fixed. The statistical albedo and \textit{virtual light stage} illumination model guide the optimization and avoid mixing  intrinsic albedo and illumination. The loss is:
\begin{align}
\label{eq:loss1}
    \operatorname*{argmin}_{(\mathsf\omega, \mathsf\alpha, \mathsf\delta, \mathsf\beta, \mathsf\gamma, \mathsf{R}, \mathsf{T})} \mathsf{E}_{d}(\chi) + \;\mathsf{E}_{p}(\mathsf\alpha, \mathsf\beta, \mathsf\gamma, \mathsf\omega) + \mathsf{E}_{b}(\mathsf\gamma, \mathsf\delta)
\end{align}
\begin{figure*}
    \includegraphics[width=\linewidth]{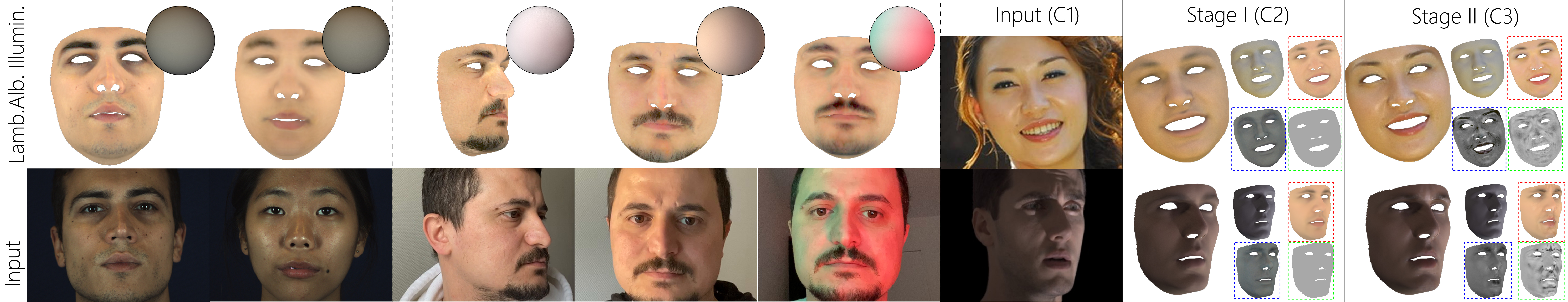}
  \caption{Left: Consistency of the estimated light for different subjects under the same lighting condition. Right: Consistency of the estimated diffuse albedo for the same subject under different lighting conditions and poses. Right: Stage II, C2 refines the estimated priors of Stage I, C1 and capture person specific facial details in the final diffuse (red), specular (blue), roughness (green) albedos. Even under strong directional light (second row), our method successfully captures the shadows and produces shadows-free personalized albedos.}
  \label{fig:consistency}
  \vspace{-20px}
\end{figure*}
With $\mathsf{E}_{d}(\chi) = \mathsf{E}_{ph}(\chi) + \alpha_1\;\mathsf{E}_{land}(\chi)$ and $\mathsf{E}_{p}(\mathsf\alpha, \mathsf\beta, \mathsf\gamma, \mathsf\omega)$ is a prior that ensures optimization tractability and given by $\mathsf{E}_{p}(\mathsf\alpha, \mathsf\beta, \mathsf\gamma) + w_1\; \mathsf{E}_{p}(\mathsf\omega)$. $\mathsf{E}_{p}(\mathsf\alpha, \mathsf\beta, \mathsf\gamma)$ is the statistical face (shape and albedo) prior that regularizes against implausible face geometry and reflectance deformations, and given by $\mathsf{E}_{p}(\mathsf\alpha, \mathsf\beta, \mathsf\gamma) =  w_i \sum_{k = 1}^{K_s} (\frac{\mathsf\alpha_k}{\sigma_{\mathsf\alpha_k}}) + w_c \sum_{k = 1}^{K_r} (\frac{\mathsf\beta_k}{\sigma_{\mathsf\beta_k}}) + w_s \sum_{k = 1}^{K_s} (\frac{\mathsf\gamma_k}{\sigma_{\mathsf\gamma_k}})$. $\sigma_{\mathsf\alpha_k}$, $\sigma_{\mathsf\beta_k}$ and $\sigma_{\mathsf\gamma_k}$ are the standard deviations for shape, diffuse and specular albedo, respectively. Light intensity regularizer $\mathsf{E}_{p}(\mathsf\omega) =\sum_{j=0}^{M}||\mathsf{I}_j-\mathsf{m}_j||_2^2$, where $\mathsf{m}_j$ is mean intensity of the $j^{th}$ light. We observe that the final illumination is sensitive to weight $w_1$, where high value for $w_1$ leads to monochromatic illumination, while smaller values favor multi-colored illumination. For all our experiments, we use $w_1 = 0.01$, that helps model various illumination scenarios and avoids baking albedos into illumination. Finally, $\mathsf{E}_{b}(\mathsf\delta, \mathsf\omega)$ is a box constraint that restricts $\mathsf\delta$ to range $[0, 1]$. $\mathsf{d}_j>0$, $\mathsf{a}_j>0$, $\mathsf{i}_j>0$ and $\mathsf{p}_j$ ensure that the area lights stay within the homothetic icosahedron faces.
\vspace{4px}
\\
\textbf{Stage II.} Albedos obtained in Stage I captures the base diffuse and specular statistical albedos. 
In this stage, we capture personalized face skin attributes -- diffuse $\mathcal{\hat{C}}$, specular $\mathcal{\hat{S}}$ and roughness $\mathcal{\hat{R}}$. We use optimized $\mathcal{C}$ and $\mathcal{S}$ to initialize personalized albedos $\mathcal{\hat{C}}$, $\mathcal{ \hat{S}}$, and uniform initial roughness $\mathcal{\hat{R}}$ with loss:
\begin{align}
\label{eq:loss3}
\operatorname*{argmin}_{(\mathcal{\hat{C}, \hat{S}, \mathcal{\hat{R}}})} & \mathsf{E}_{d}(\hat{\chi}) + w_2( \mathsf{E}_{s}(\mathcal{\hat{C}}) + \mathsf{E}_{s}(\mathcal{\hat{S}})) + w_3 (\mathsf{E}_{c}(\mathcal{\hat{C}}, \mathcal{C}) + \mathsf{E}_{c}(\mathcal{\hat{S}}, \mathcal{S})) + \nonumber\\ 
& w_4( \mathsf{E}_{m}(\mathcal{\hat{C}}) +  \mathsf{E}_{m}(\mathcal{\hat{S}})+ \mathsf{E}_{m}(\mathcal{\hat{R}}))  + (\mathsf{E}_{b}(\mathcal{\hat{S}}) + \mathsf{E}_{b}(\mathcal{\hat{R}}))
\end{align}
Here, $\hat{\chi} = \{\mathsf\omega, \mathsf\alpha, \mathsf\delta, \mathcal{ \hat{C}}, \mathcal{ \hat{S}}, \mathcal{\hat{R}}, \mathsf{R}, \mathsf{T}\}$ is new parameters set and $\mathsf{E}_{b}(\mathcal{\hat{S}})$ (resp. $\mathsf{E}_{b}(\mathcal{R})$) is the soft box constraints that restrict the specular (resp. roughness) to remain in an acceptable range $[0, 1]$. $\mathsf{E}_{m}(\mathcal{\hat{C}})$ (resp. $ \mathsf{E}_{m}(\mathcal{\hat{S}})$ and $ \mathsf{E}_{m}(\mathcal{R})$) is a constraint term that ensures local smoothness of each vertex, with respect to its first ring neighbors in the UV space, and given by $\mathsf{E}_{m}(\mathcal{\hat{C}}) = \sum_{\mathsf{x}_j  \in \mathcal{N}_{\mathsf{x}_i}} ||(\mathcal{\hat{C}}(\mathsf{x}_j) - \mathcal{\hat{C}}(\mathsf{x}_i)||_2^2$, where $\mathcal{N}_{\mathsf{x}_i}$ is 4-pixel neighborhood of pixel $\mathsf{x}_i$. 
\\
$\mathsf{E}_{s}(\mathcal{\hat{C}})= \sum_{i\in M}| \mathcal{\hat{C}}(\mathsf{x}_i)) - \texttt{flip}(\mathcal{\hat{C}}(\mathsf{x}_i)))|_1$ is a symmetry constraint, where $\texttt{flip}()$ is the \textit{horizontal flip} operator, similar to \cite{tran2018nonlinear}. $\mathsf{E}_{c}(\mathcal{\hat{C}}, \mathcal{C})$ is a consistency regularizer that weakly regularizes the optimized $\mathcal{\hat{C}}$ with respect to the previously optimized statistical albedo $\mathcal{C}$ based on the chromaticity $\kappa$
of each pixel in the texture, given by, $\mathsf{E}_{c}(\mathcal{\hat{C}}, \mathcal{C} ) = \sum_{i\in M} |\kappa(\mathcal{\hat{C}}(\mathsf{x}_i))-\kappa(\mathcal{C}(\mathsf{x}_i))|_1$. $\mathsf{E}_{s}(\mathcal{\hat{C}})$ and $\mathsf{E}_{c}(\mathcal{\hat{C}}, \mathcal{C})$ help prevent residual self-shadows or specular reflections to bake into the diffuse albedo (same reasoning applies for $\mathsf{E}_{s}(\mathcal{\hat{S}})$ and $\mathsf{E}_{c}(\mathcal{\hat{S}}, \mathcal{S} )$).
\vspace{4px}
\\
Intuitively, when the side of the face is under a shadow, the estimated shadow due to illumination approximation (Stage I), may not fully estimate the real shadow in the input image, while Equation~\ref{eq:loss3} tries to extract meaningful information from the image. Thus, a residual shadow, not fully estimated due to illumination approximation, can get baked into $\mathcal{\hat{C}}$. $E_{s}(\mathcal{\hat{C}})$ is a symmetric regularizer that prevents baking of the residual shadow into $\mathcal{\hat{C}}$, penalizing for a image-based imbalance between the two sides of the face. $E_{c}(\mathcal{\hat{C}}, \mathcal{C})$ the consistency regularizer, makes sure that diffuse albedo is closer to the statistical diffuse albedo, than the self-shadow's chromaticity.
\\
We note that although the method can be iterated over the Stage I and II, this iteration did not provide substantial improvements in the final results or refinements in disentangling.
\vspace{4px}
\\
\textbf{Edge Sampling.} An important limitation of differentiable ray tracing is the discontinuities present around geometric edges. That is, when solving for the rendering equation \cite{Kajiya86Rendering} via Monte Carlo ray tracing, very few points on the edge of the geometric shape are sampled, causing a discontinuity along the edges. As a result, back-propagation based gradients calculation fails to take into account sensitive information along the geometric edges. Consequently, the gradients on the edges remain \textit{noisy}, and optimization does not use the \textit{true} gradient during an iteration, especially while optimizing for affine transformations and geometric shape change.
\\
One solution is to use high number of sample points for sampling along edges. However, this is computationally infeasible. Several techniques \cite{Loubet2019Reparameterizing,Li:2018:DMC} have been proposed to overcome this limitation. In our work, we rely on \cite{Li:2018:DMC}'s technique to explicitly sample the geometry edges -- a costly yet mandatory operation needed for correct geometric shape estimation.
\vspace{4px}
\\
\textbf{Variance Reduction.} Another aspect when using differentiable ray tracing is image variance due to Monte-Carlo random sampling. Choosing an appropriate sampling strategy can drastically reduce this variance. While, a naive increase in the number of samples can reduce the variance, it is computationally expensive. We use importance sampling \cite{pharr2016physically,Li:2018:DMC} with 16 samples/pixel and then apply Gaussian smoothing over the synthesized image with a kernel of size $3\times 3$ and $\sigma=1$. Due to this smoothing operation, variance is considerably attenuated and optimization converges faster.

%% file: 5.results.tex
\section{Results and Implementation}
\label{sec:results}
\begin{figure*}
  \includegraphics[width=\linewidth]{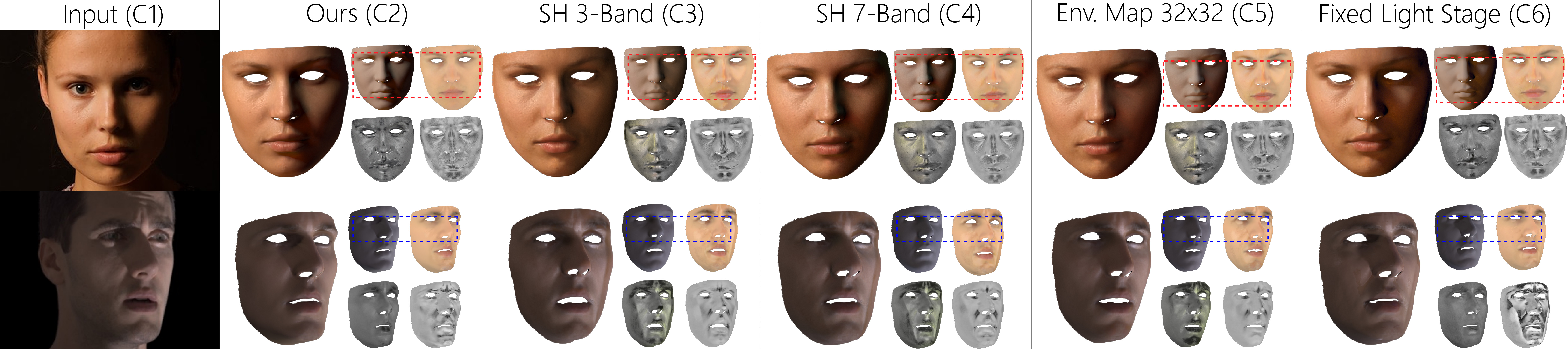}
  \caption{From left to right. C1: Input image C2: Results obtained by our method C3: Results obtained by using spherical harmonics (SH) 3-bands C4: Results obtained with SH 7-bands C5: Reconstruction using an environment map C6 : Reconstruction using a fixed light stage}
  \label{fig:ablation}
  \vspace{-20px}
\end{figure*}
We created a dataset of images with various illuminations, self-shadows (hard and soft), ethnicity, facial hair, skin types, expressions and poses to assess the robustness and quality of the reconstruction, Figure~\ref{fig:results}. For each subject, we show the final reconstruction, along with the estimated reflectance (diffuse, specular and roughness), estimated illumination and self-shadows. 
Subjects 1-5 (Fig~\ref{fig:results}) and $2^{nd} \text{and}\;3^{rd}$ subjects in Fig~\ref{fig:teaser} shows disentangled attributes of neutral face shapes, expressions, shadow-free albedos and light directions, under challenging lighting conditions.
\\
For Subject 1, the optimized light produces sharp shadows, true to the input image. Subjects 3 and 5, show reconstruction from gray scale input images. Here, a blueish light estimate compensates for the red and yellow components and produces a final gray-scale result similar to the input image, and a meaningful diffuse albedo is also reconstructed (similarly for Subject 12). In addition to handling hard shadows, we show in Subjects 6-8, the ability to produce \textit{soft} shadows. For Subject 6, we get a fair reconstruction under a directional light. Subjects 9-11, 15 have visible specular areas on their faces. Our method successfully extracts specular highlights seen in specular and roughness reconstructions. 
\\
Subjects 12-17 show reconstructions for people with various skin pigmentations, colorations, facial hair and ethnicities. Our method captures person specific details in the optimized diffuse albedo. Subject 18 (from ~\cite{zhang2020portrait}), with challenging lighting conditions is shown, where the face is lit by incoming light from the bottom right~\footnote{See supplementary video for shadow edition results.} and a hard shadow on the subject's nose. The estimated light captures this shadow and produces shadow-free albedos. Subject 19 is a failure case from~\cite{zhang2020portrait}, our method provides a good estimate of self-shadows (especially under the eyes). 
\vspace{4px}
\\
\textbf{Implementation Details.} Our framework is implemented using PyTorch~\cite{paszke2017automatic} with a GPU enabled backend (NVIDIA GeForce RTX 2080 GPU and Intel i7 9800X). Ray tracing is based on the method of \cite{Li:2018:DMC}, and for optimization we use Adam~\cite{kingma2014adam} with default $\beta_1=0.9, \beta_2=0.999$ and $\alpha_1 = 1$. In-order to weight all parameters equally during the optimization, we use different learning rates (\texttt{lr}) for each parameter. For light stage parameters we use $\texttt{lr}=0.001$, for statistical albedo $\texttt{lr}=0.02$ and for shape identity $\texttt{lr}=0.01$. Camera rotation, translation and blendshapes use $\texttt{lr}=0.001$. Finally for the diffuse, specular and roughness, we use $\texttt{lr}=0.005$. For regularization we use $w_i = 0.0025$,  $w_c = w_s = 0.0025$, $w_1 = 0.01$, $w_2 = w_3 = 0.3$ and $w_4 = 0.0002$. The processing time of our method depends on input image resolution. An image of resolution $512\times 512$ takes about 6.4 minutes (wall-clock time) for the full optimization, where Stage I takes 5.1 minutes and Stage II takes 1.3 minutes.

%% file: 6.ablation.tex
\section{Ablation Studies}
\label{ssec:ablation}
We show ablation studies on comparison against fixed light stage and the importance of the Stage II to capture personalized skin reflectance. We refer the reader to the supplementary material (section  B) for additional ablation studies on the choice of geometries for the light stage. 
\vspace{3px}
\\
We validate the importance of our parameterized virtual light stage. A fixed light stage is created, where the light intensity $\mathsf{I}_j$ is now a parameter -- not dependent on $\mathsf{d}_j$ or $\mathsf{a}_j$ -- fully unconstrained. The light surface-area and position are fixed and not optimized and only the light intensity is optimized. We observe that this optimization formulation gives less accurate shadow estimation and leads to suboptimal light-albedo disentangling (Figure \ref{fig:ablation}, C6). Adding structure to $\mathsf{I}_j$ parameterization (Equation~\ref{eq:lightIntensity}) leads to substantially better results as shown on Figure \ref{fig:ablation}, C2. Figure~\ref{fig:consistency} (left), discusses the effectiveness of Stage II personalization to refine over Stage I's result. Figure~\ref{fig:consistency} shows the consistency of the estimated light and albedos under various input image and subject conditions.

%% file: 6.comparisons.tex
\section{Comparisons}
\label{sec:comparisons}
\begin{figure*}
\includegraphics[width=\linewidth]{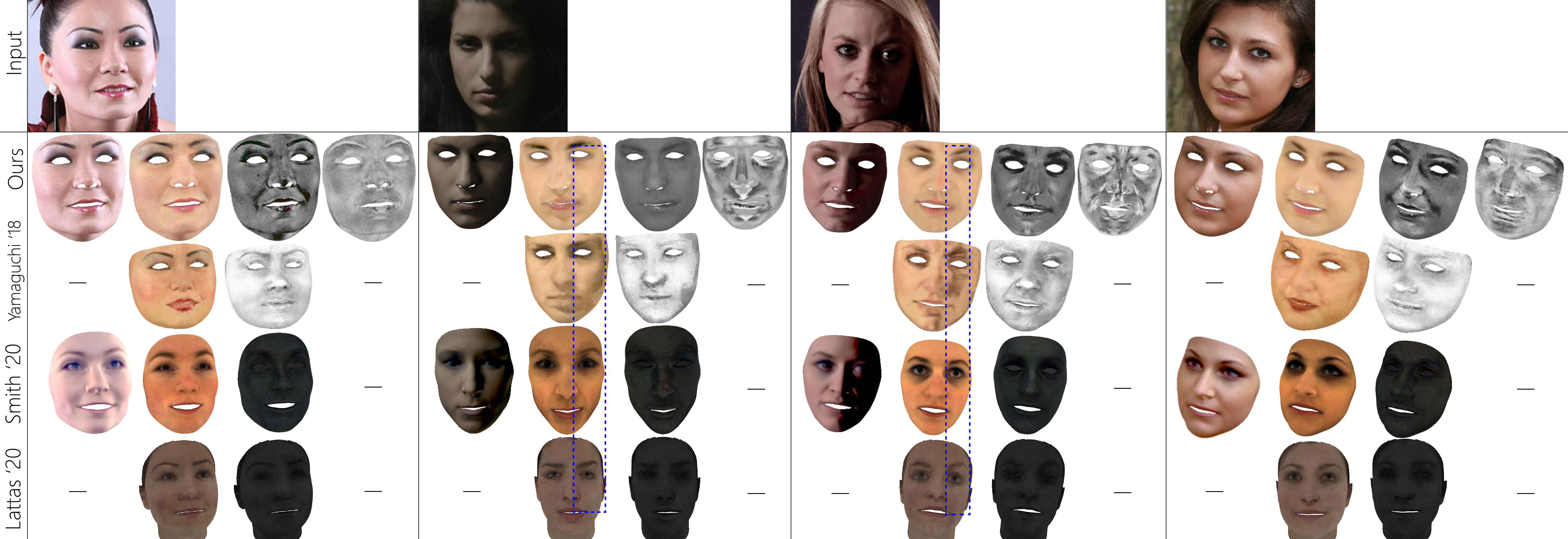}
  \caption{\eg{For each subject (left to right), we compare final reconstruction, diffuse, specular, and roughness albedos with \cite{yamaguchi2018high,smith2020morphable, lattas2020avatarme}. \cite{yamaguchi2018high, lattas2020avatarme} final reconstruction is not available as their method do not estimate scene light; none of the other methods explicit estimate roughness.}}
  \label{fig:compResDeep}
  \vspace{-20px}
\end{figure*}
\begin{figure*}
    \includegraphics[width=\linewidth]{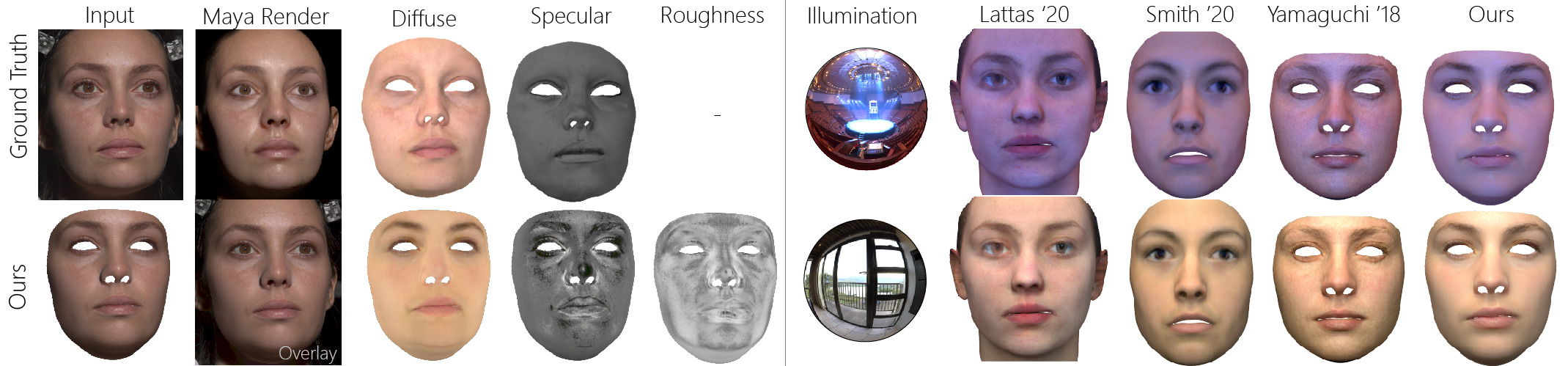}
  \caption{\eg{Left: Comparison of our method with Ground Truth (GT) data of the Digital Emily project. Right: Relighting comparison.}}
  \label{fig:emily}
  \vspace{-25px}
\end{figure*}

\textbf{Geometry and Albedo.} We compared the geometric reconstruction error against state-of-the-art methods, \cite{tewari17MoFA}, \cite{tran2019towards}, \cite{chen2019photo}, and \cite{lattas2020avatarme}, where twenty four ground truth geometries from \cite{3ddfa_cleardusk, Pillai2019_3dfaw} are used. Our method outperforms these methods and the results are available in the supplementary (Section D). We also compare against state-of-the-art methods \cite{yamaguchi2018high}, \cite{smith2020morphable}\footnote{Using \href{https://github.com/waps101/AlbedoMM}{https://github.com/waps101/AlbedoMM}} and \cite{lattas2020avatarme}, that extract both diffuse and specular albedos (Figure~\ref{fig:compResDeep}). Note that methods \cite{yamaguchi2018high} and \cite{lattas2020avatarme} does not model scene illumination and directly infer skin reflectance attributes, so we do not have their final image render. 
For the same reason, without given illumination, 
their methods can bake some self-shadow information into the estimated diffuse and specular albedos, as highlighted (in blue) in Figure~\ref{fig:compResDeep}. 
\\
\eg{We note that \cite{yamaguchi2018high} and \cite{lattas2020avatarme} estimates displacement/normal maps while our method does not. This requires high-quality and well lit input images (as reported by authors) for optimal results. Additionally, \cite{lattas2020avatarme} estimates reflectance maps for full face head in the UV space, whereas our method restricts reconstruction to frontal face only}. \cite{smith2020morphable} estimates light (three bands spherical harmonics) but, \eg{may not correctly estimate personalized reflectance outside the statistical albedo space.} A complete catalog of comparisons against these methods is available in the supplementary material (section C). Additionally, we also compare our method with \cite{tewari17MoFA,tran2019towards,sengupta2018sfsnet}, see supplementary (Section C).
\vspace{4px}
\\
\textbf{Digital Emily.} In Figure~\ref{fig:emily}, we compare our method with the ground truth (GT) data 
from the Digital Emily~\cite{emilyWikihuman} project. In addition, we compare quantitatively, our image reconstruction quality against state-of-the-art (see Table~\ref{tab:emilyDiffSpec}). For each method, we compute SSIM (max: 1.0) and PSNR (dB) scores for final render, Ground-Truth (GT) diffuse, and GT specular image pairs (GT roughness not compared due to unavailability). Each image is rendered from the GT camera space using a mask depicted in Figure~\ref{fig:emily} (bottom-left). As shown in Table~\ref{tab:emilyDiffSpec}, \eg{our method provides images with the highest similarities in SSIM for diffuse rendered image. For PSNR (diffuse, specular) and SSIM (specular) \cite{yamaguchi2018high} scores slightly better than our method.} \eg{Please note that since each method has a different UV map parametrization, we did the comparison on the face mask image renders and not on unwrapped texture space}. As \cite{yamaguchi2018high} and \cite{lattas2020avatarme} do not estimate scene light, so we do not have comparison of the final image renders against GT. Finally, we compare rendered GT images (using Autodesk Maya) against input image and obtain $SSIM = 0.973$, $PSNR = 36.526$. We note that our final image render vs. input image have scores $SSIM = 0.982$, $PSNR = 41.475$ that are closer to the input image.
\vspace{4px}
\\
\begin{table}
\begin{adjustbox}{max width=\linewidth}
\centering
\begin{tabular}{|lllllll|} 
\hline
{vs GT~} & {Final~} & {Final~} & {Diffuse~}       & {Diffuse~} & {Spec.~} & {Spec.}   \\
{Render} & {(SSIM)} & {(PSNR)} & {(SSIM)}         & {(PSNR)}   & {(SSIM)} & ({PSNR)}  \\ 
\hline
{Ours}    & {\textbf{0.965}}  & {\textbf{36.390}} & {\eg{\textbf{0.722}}} & \eg{29.812}                                 & \eg{0.547}                       & \eg{29.670}                                \\
{\cite{yamaguchi2018high}}   & -                                    & -                                    & \eg{0.679}                                        & \eg{\textbf{30.061}}                        & \eg{\textbf{0.604}}                                & \eg{\textbf{30.923}}                       \\
{\cite{smith2020morphable}}  & 0.906                                    & 35.389                                    & \eg{0.639}                                        & \eg{29.006}                                 & \eg{0.452}                                & \eg{28.833}                                \\
{\cite{lattas2020avatarme}} & -                                    & -                                    & \eg{0.540}                                        & \eg{28.633}                                 & \eg{0.516}                                & \eg{28.926}                                \\
\hline
\end{tabular}
\end{adjustbox}
 \caption{Final, diffuse and specular albedos in comparison with GT Maya renders for our, \cite{yamaguchi2018high}, \cite{smith2020morphable} and \cite{lattas2020avatarme}. SSIM and PSNR (dB): higher the better.}
 \label{tab:emilyDiffSpec}
 \vspace{-20px}
\end{table}
\textbf{Spherical Harmonics (SH) vs. Light Stage.} In this experiment, we use Spherical Harmonics (SH) to model light instead of the light stage (Figure \ref{fig:ablation}). First subject (first row Fig~\ref{fig:ablation}), three-bands SH (C3) provides a coarse estimation of the light, and the shadow is barely captured, where estimated albedos get some light and shadows baked into it. Seven-bands SH (C4) captures more shadows but still produces sub-optimal disentangling vs. our light stage (C2). For the second subject (second row), the hard shadow cast by the nose was only captured by our light stage while (3 and 7 bands) SH are visually inaccurate. We also experimented with higher-order SH band (9 and 11) without substantial improvements, especially for subject in row two,  Fig~\ref{fig:ablation}. These experiments shows that using high-order SH can be used to obtain meaningful shadows estimations, but fails to capture hard shadows produced by point lights in the scene, and leads to sub-optimal disentangling. 
Finally, our parametric light stage models semantically meaningful light types -- point, directional, while basis functions used by SH only model lights at infinity and are harder to manipulate intuitively (e.g. for shadow removal applications).
\vspace{4px}
\\
\textbf{Environment Map vs. Light Stage.} In this experiment, we replaced the light stage with an environment map to model lighting. Each pixel in the environment map , $32\times 32$ resolution, represents a light source at infinity, where light intensity of each pixel is parameterized. Results for this optimization are shown in Figure \ref{fig:ablation} (C5). Because environment map can only model lights at infinity, is not flexible enough to model arbitrary (e.g area) lights, opposed to the lightstage, and thus, fails to capture the shadows generated by point lights (for both subjects) and produces sub-optimal disentangling.

%% file: 7.applications.tex
\section{Applications}
\label{sec:applications}
Robust estimation of reflectance and illumination provides explicit control over these attributes, with several practical applications: relighting, light transfer, shadow and specularity editing, and image texture filters addition.
\vspace{2px}
\\
\textit{Illumination Edition and Transfer.}
\label{sssec:relighting}
Figure~\ref{fig:teaser} (right) first column shows relighting under novel illumination conditions. Second column, shows results for estimated light transfer, where estimated light from source image is used to illuminate target subject. Source image's self-shadows, due to illumination, are successfully transferred in the target render.
\vspace{2px}
\\
\textit{Shadow and Flash Removal.} \footnote[6]{The reader is referred to supplementary video for better visualization} Inspired by \cite{zhang2020portrait}, we show self-shadow removal application. While, \cite{zhang2020portrait}'s method can remove shadows cast from external (foreign) objects; our method handles self-shadow removal, as shown in Figure~\ref{fig:application2} (left). In the accompanying video, we also show demonstration of camera flash removal for face images, where estimated illumination from first image replaces estimated illuminations in subsequent image frames.
\vspace{2px}
\\
\textit{Albedo Edition and Transfer.} Third column in Figure \ref{fig:teaser} (right) shows diffuse and specular albedo transfer applications, from thumbnail source to target image, while the last column shows the result of applying a texture filter (using multiplication operator) on the optimized diffuse albedo in the UV space. Finally, in Figure~\ref{fig:application2}\footnotemark[6] (right), we show an application where estimated specular albedos can be edited on portrait images. This is done by gradually decreasing the estimated roughness, while using a constant estimated base specular albedo. 
\vspace{-3px}

%% file: 8.conclusion.tex
\section{Limitations and future works}
\label{sec:limitations}
\begin{figure}[t]
  \includegraphics[width=\linewidth]{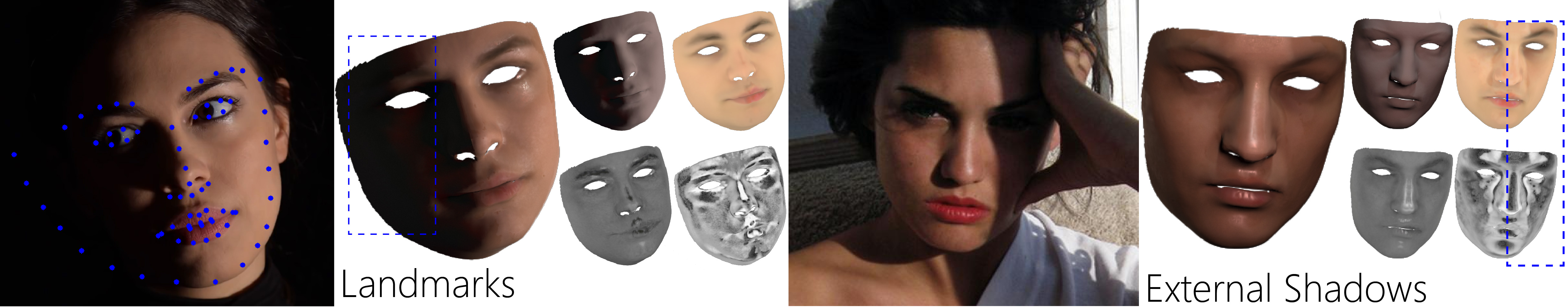}
  \caption{Limitations -- left: Imprecise landmarks under extremely scene illumination produces incorrect geometry reconstruction. Right: External shadows get baked into albedos.}
  \label{fig:failure}
   \vspace{-25px}
\end{figure}

\vspace{-3px}
\textbf{Limitations.} Our method relies on sparse landmarks for pose and geometry estimation. While this works well for several illumination scenarios, in extreme partial darkness (Figure \ref{fig:failure}, left), landmarks estimates and subsequent geometry estimation are less accurate. In several such cases, human landmarks estimation can also be incorrect, thus, a better approach to handle such cases is needed. Our method does not model external shadows (Fig~\ref{fig:failure}, right), in that case our method could benefit from a method such as~\cite{zhang2020portrait}. \eg{Another limitation of our method is reliance on statistical albedo priors (Optimization, Stage I) that do not model certain skin tones. As a result, non-Caucasian albedos may not be estimated correctly. The unexplained diffuse albedo can get baked into the illumination, especially for darker skin tones, as shown in Figure \ref{fig:results}, Subject 20.}
\vspace{3px}
\\
\eg{We note that our albedos (esp. roughness) attributes are view and input image illumination condition dependent, however, when available, statistical priors help give meaningful estimates. Here, our method relies on symmetry, consistency and smoothness regularizers (Eq~\ref{eq:loss3}) to avoid overfitting. In some cases, due to these regularizers, person specific attributes are not captured. Additionally, while the consistency and symmetry regularizers (Stage-II) help avoid baking shadows in the final albedo, in some cases, when the optimized light and consequent shadows are inaccurate, some light/shadow patches may appear in the estimated albedos. Finally, the proposed light stage may not always recover accurate illumination for certain illumination conditions. For instance, because we model a single area light per icosahedron face, in case there are several light sources in one direction, the light stage may either favor the main light in this direction or an average of these lights.}
\\
\begin{figure}
  \includegraphics[width=\linewidth]{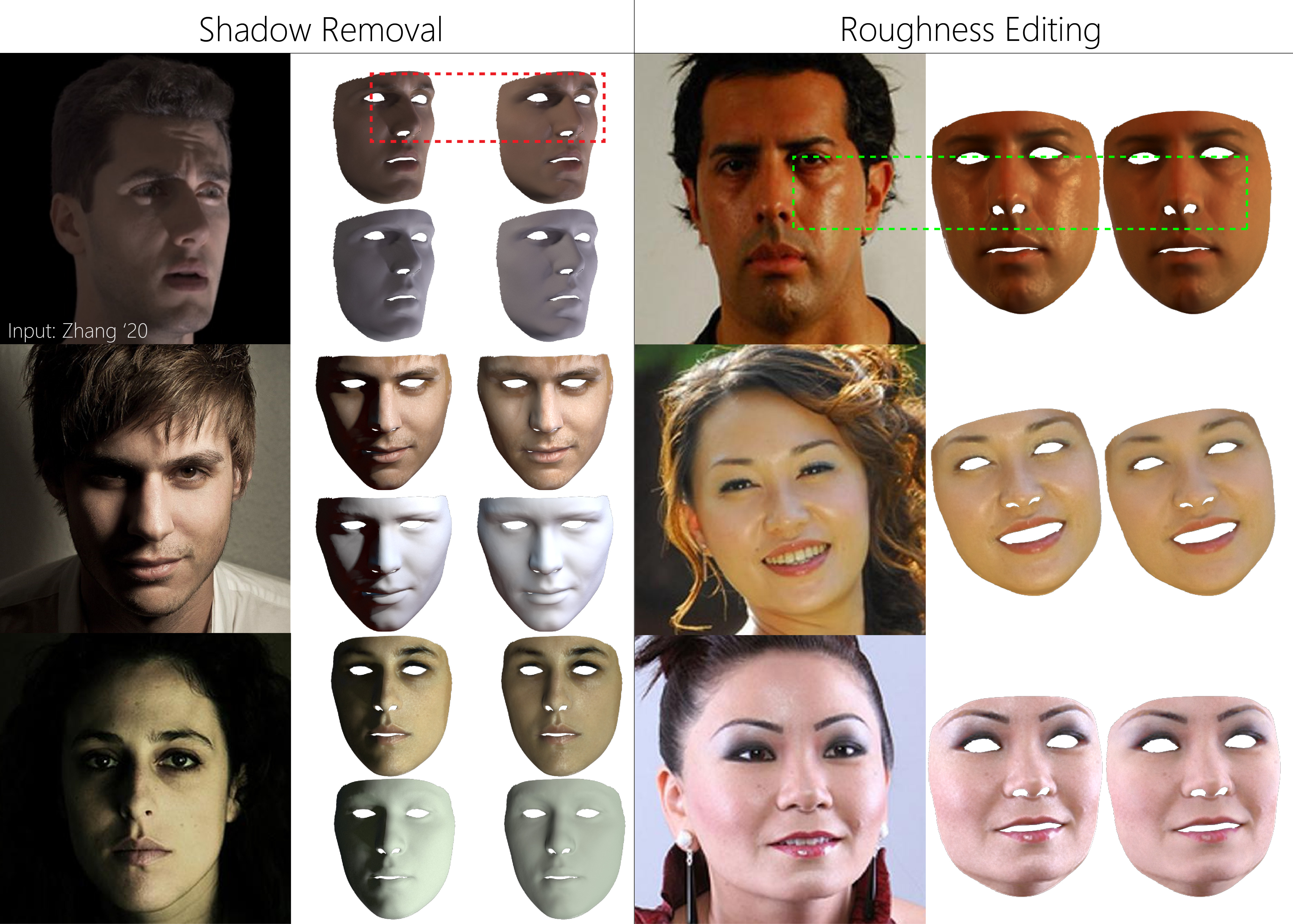}
  \caption{Left: We show self-shadow editing removing resultant self-shadows (in red) by manipulating optimized illumination to uniform illumination. Right (Input, optimized, and edited specular highlights): By manipulating the optimized roughness map, specular reflections (in green) can be edited.}
  \label{fig:application2}
  \vspace{-15px}
\end{figure}
\textbf{Future Works.} In the future, we want to extend our approach with methods such as \cite{li2020learning}, to model mesoscopic geometric details, \cite{yamaguchi2018high}. Currently, we use single bounce rays for illumination modeling due to lack of external scene geometries, a natural extension is to model multi-ray bounces for softer shadows. Further, our methods naturally extends to a multi-view face reconstruction formulation that would help improve attribute estimation quality. Finally, we plan to extend our method with more complex skin reflectance models such as BSSRDF/dielectric materials, \cite{weyrich2006analysis}. 
\vspace{3px}
\\
\vspace{-20px}
\section{Conclusion}
\label{sec:conclusion}
We present a novel and robust face modeling approach, under general illumination conditions. A virtual light stage formulation to model scene illumination is introduced, which, used in-conjunction with a differentiable ray tracing, makes our method self-shadows and specular reflectance aware. We then formulate face modeling as a loss minimization problem, and solve it via a two-stage optimization strategy. This strategy systematically disentangles face attributes, that make the optimization tractable for unconstrained input images. To validate our method, along with several results, we provide ablation studies, analysis of various modeling decisions and limitations. 
Beyond its accuracy and robustness to light conditions, the  rich decomposition resulting from our approach allows for several style -- illumination and albedo -- transfer and edit applications. 
\footnotemark[0]{
\textbf{Acknowledgements.} We thank the anonymous reviewers for their feedback. A sincere thank you to Christel Chamaret and Lionel Oisel for their support.}

%% file: 9.appendix.tex
\section{Face Reflectance Model}
\label{appendix:Cook-Torrance}
We provide the interested reader the Cook-Torrance BRDF used to model the human face reflectance. Equations presented here are based on~\cite{walter2007Microfacet}. The Cook-Torrance BRDF models a surface as small micro-facets where each facet interacts differently with the incoming light depending on its local normal, roughness and specular parameters. The BRDF is represented as:
 \begin{align}
 f_r(\mathsf{s}_i,r_i, \mathsf{c}_i, \mathsf{n}_i, \mathsf{l}, \mathsf{o}) &= f_{d}(\mathsf{c}_i) + f_{s}(\mathsf{s}_i, r_i, \mathsf{n}_i, \mathsf{l}, \mathsf{o})
 \end{align}
 where, $\mathsf{c}_i$ is the intrinsic albedo color of the surface point. $\mathsf{s}_i \in \mathbb{R}^3$ represents the base reflectivity of a point. $r_i \in \mathbb{R}$ is the roughness term, $\mathsf{n}_i$ is the normal vector at point $\mathsf{v}_i$, $\mathsf{l}\in\mathbb{R}^3$ is the light direction and $\mathsf{o}\in\mathbb{R}^3$ is the view vector. $f_{d}$ is the Lambertian diffuse term:
\begin{equation}
 \label{eq:brdf_diffuse}
 f_{d}(\mathsf{c}_i) = \frac{\mathsf{c}_i}{\pi}
 \end{equation}
and $f_{s}$ the specular and roughness ($r_i$) term: 
\begin{equation}
 \label{eq:brdf_spec}
 f_{s}(\mathsf{s}_i, r_i, \mathsf{n}_i, \mathsf{l}, \mathsf{o}) = \frac{D(\mathsf{h}, \mathsf{n}_i, r_i) * G(\mathsf{h}, \mathsf{n}_i, r_i) * F( \mathsf{s}_i, \mathsf{o},  \mathsf{l}) }{4 *  (\mathsf{o} \cdot \mathsf{n}_i)(\mathsf{l} \cdot \mathsf{n}_i)  }
 \end{equation}
$D$ is the statistical distribution of surface normal over the micro-surface. In our case, the Blinn-Phong distribution is used, given by:
 \begin{equation}
 \label{eq:brdf_d}
 D(\mathsf{h}, \mathsf{n}_i, r_i) = \frac{\frac{2}{r_i} + 2 }{2 * \pi} (\mathsf{n}_i * \mathsf{h})^{ \frac{2}{r_i} - 2}
 \end{equation}
where $\mathsf{h}$ is the normalized halfway vector between $\mathsf{l}$ and $\mathsf{o}$:
 \begin{equation}
 \label{eq:halfVector}
 \mathsf{h} = \frac{\mathsf{l} + \mathsf{o}}{\left \| \mathsf{l} \cdot \mathsf{o} \right \| }
 \end{equation}
$G$ is the bidirectional shadow masking function that describes which portion of the micro-facet is under a shadow in both directions $\mathsf{o}$ and $\mathsf{l}$. In our case, Smith~\cite{smith1967Geometrical} shadowing masking approximation is used, that approximates $G$ as product of two mono-directional shadowing terms: 
  \begin{equation}
 \label{eq:brdf_g}
 G(\mathsf{h}, \mathsf{n}_i, r_i, \mathsf{o}, \mathsf{l}) \approx G_1(\mathsf{n}_i, \mathsf{o}, r_i) \cdot G_1(\mathsf{n}_i, \mathsf{l}, r_i)
 \end{equation}
with $G_1$ equals to (using \cite{walter2007Microfacet} approximation) : 
\begin{equation}
 \label{eq:brdf_g1}
  G_1(\mathsf{n}_i, \mathsf{v}, r_i)= \Delta(\frac{\mathsf{h} \cdot \mathsf{v}}{\mathsf{n}_i \cdot \mathsf{v}} ) \cdot \begin{cases}
    \frac{3.535 a + 2.181 a^2}{1 + 2.276 a + 2.577 a^2}, & \text{if $a<1.6$}.\\
    1, & \text{otherwise},
  \end{cases}
\end{equation}
with $a$ equal to: 
 \begin{equation}
 \label{eq:brdf_g1_a}
  a = \frac{1}{r_i \cdot tan(cos^{-1}(\mathsf{n}_i \cdot \mathsf{v})}
\end{equation}
and 
\begin{equation}
\label{eq:brdf_g1_chi}
  \Delta (x)=\begin{cases}
    1, & \text{if $x>0$}.\\
    0, & \text{otherwise}
  \end{cases}.
\end{equation}

Finally, $F$ is  the Fresnel term that describes the amount of reflected light in a given direction. We use Schlick approximation  \cite{Schlick94aninexpensive} for $F$ equal to: 
\begin{equation}
 \label{eq:brdf_f}
  F( \mathsf{s}_i, \mathsf{o}, \mathsf{l}) = \mathsf{s}_i + (1 - \mathsf{s}_i) \cdot ( 1 - \mathsf{o} \cdot \mathsf{h})
\end{equation}
\input{10.abblation-continued}
\section{Face Catalogs}
\label{appendix:faceCatalog}
\begin{figure*}
\includegraphics[width=\linewidth]{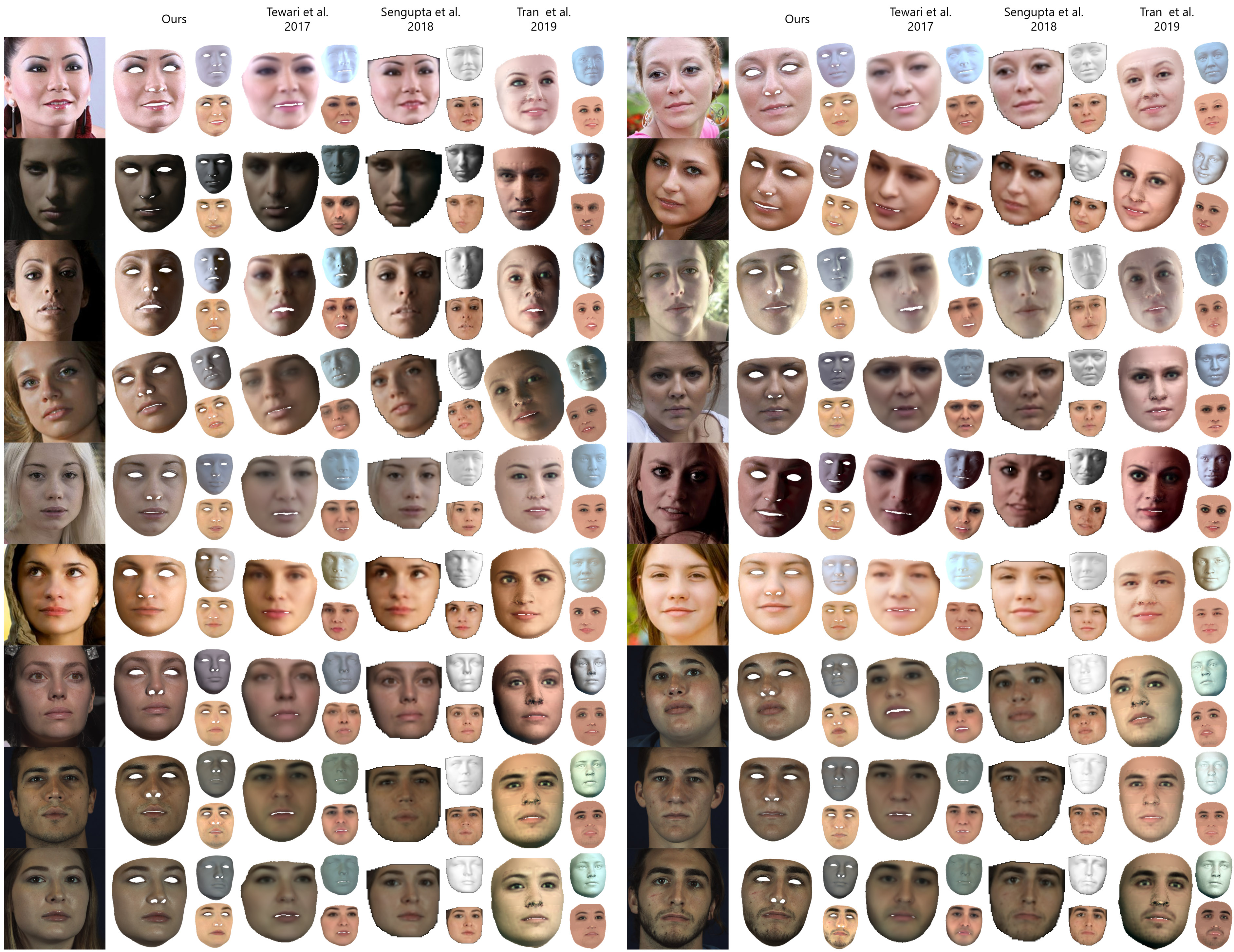}
  \caption{Examples of the final, illumination, and diffuse triplets compared to \cite{tewari17MoFA}, \cite{sengupta2018sfsnet}, and \cite{tran2019towards}.}
  \label{fig:faceCatalog1}
\end{figure*}

On Figure \ref{fig:faceCatalog1}, we show comparison of our reconstruction and estimated face and light parameters with those obtained from  \cite{tewari17MoFA,tran2019towards}\footnote[1]{Results obtained from authors} and \cite{sengupta2018sfsnet}\footnote[2]{Using \href{https://github.com/senguptaumd/SfSNet}{https://github.com/senguptaumd/SfSNet}}. These methods rely on Lambertian reflectance model combined with spherical harmonics (SH) illumination, neither model self-shadows nor specular reflections. So, we compare their SH illumination with our virtual light stage illumination and their diffuse albedo with our estimated diffuse albedo $\mathcal{\hat{C}}$. Only \cite{tran2019towards} estimates a personalized diffuse albedo similar to our, while \cite{tewari17MoFA} and \cite{sengupta2018sfsnet} only estimate statistical prior-based diffuse albedos, Figure~\ref{fig:faceCatalog1}. Our reconstruction is self-shadows and specularity aware, and avoids baking these attributes into the diffuse albedo. 
\\
\begin{figure*}
  \includegraphics[width=\linewidth]{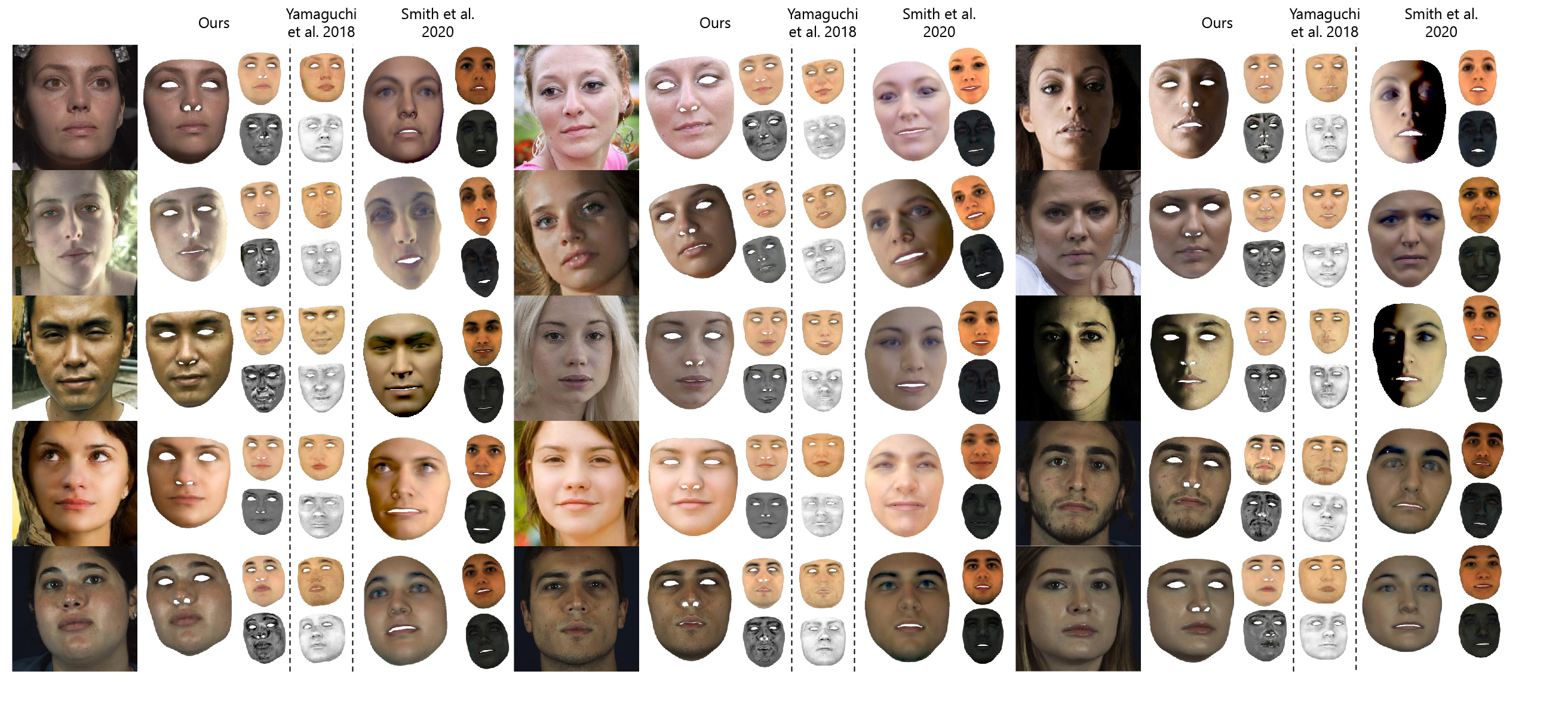}
  \caption{\eg{Examples of the final, diffuse, and specular triplets compared to \cite{yamaguchi2018high} and \cite{smith2020morphable}.}}
  \label{fig:faceCatalog2}
\end{figure*}
In Figure~\ref{fig:faceCatalog2}, we show reconstruction results compared to  \cite{yamaguchi2018high} and \cite{smith2020morphable}. We note again that, \cite{yamaguchi2018high} directly regress diffuse and specular albedos from input image and does not estimate the scene light, so the final reconstruction from their method are unavailable.
\section{Geometry Reconstruction Comparisons}
\begin{figure*}
  \includegraphics[width=\linewidth]{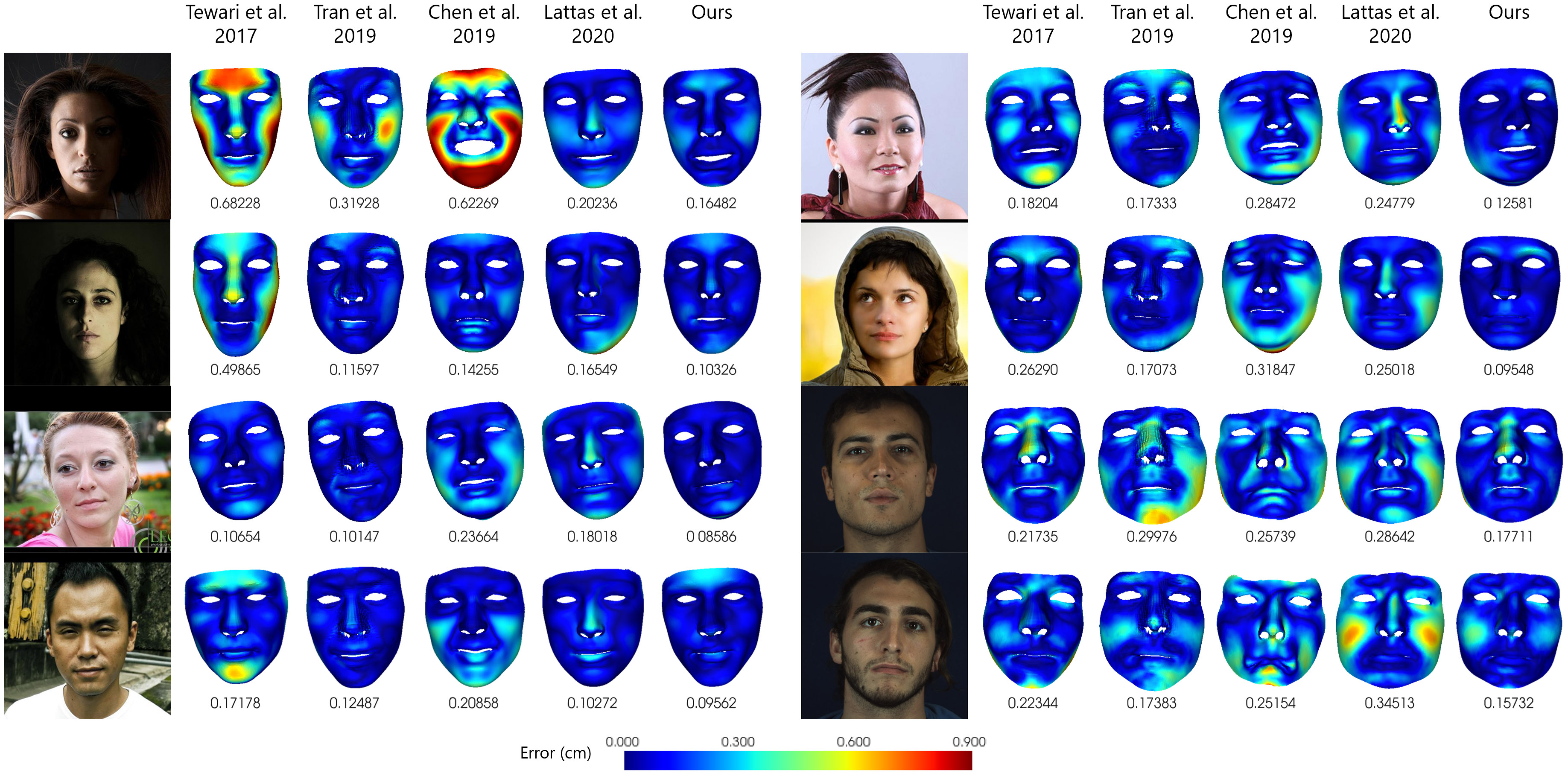}
  \caption{Quantitative evaluation of 3D mesh reconstructions. For each subject image, the first four columns show the vertex error of state-of-the-art methods, while the last column shows our results. The vertex errors are color-coded from blue to red, where vertices closer to the Ground Truth (GT) are more blue.}
  \label{fig:meshEval}
\end{figure*}

Figure~\ref{fig:meshEval} shows geometry reconstruction error against state-of-the-art methods \cite{tewari17MoFA}, \cite{tran2019towards}, \cite{chen2019photo}, and \cite{lattas2020avatarme} \footnote{Reconstruction geometries were obtained from the authors except \cite{chen2019photo}}. Vertex error was evaluated for reconstructed GT geometries provided by \cite{3ddfa_cleardusk} and \cite{zhu2017face}. This 3D face dataset is based on the AFLW dataset~\cite{koestinger2011aflw}. The last two images in the second column shown in Figure \ref{fig:meshEval} are obtained from the 3DFAW database~\cite{Pillai2019_3dfaw}, that also provides GT geometries. For a fair comparison we align all meshes to GT (see Section \ref{sec:meshDiff} for details). Comparisons show that our method has lower vertex error compared to others, especially under low light, specular or self-shadow conditions. 
\\
\begin{table}
\centering
\begin{tabular}{|l|ll|ll|} 
\hline
Distances          & \multicolumn{2}{l|}{~ ~ ~ ~ ~Position} & \multicolumn{2}{l|}{~ ~ ~ ~ ~ Normal}  \\
Method             & Mean              & Stdev              & Mean              & Stdev              \\ 
\hline
\cite{tewari17MoFA}       & 0.299260          & \eg{0.138}              & 0.187251          & \eg{0.050}              \\
\cite{yamaguchi2018high}  & 0.254348          & \eg{0.116}              & 0.155058          & \eg{0.053}              \\
\cite{tran2019towards}    & 0.173339          & \eg{0.074}              & 0.160323          & \eg{0.047}              \\
\cite{chen2019photo}      & 0.290367          & \eg{0.119}              & 0.201893          & \eg{0.057}              \\
\cite{lattas2020avatarme} & 0.201139          & \eg{0.080}              & 0.159559          & \eg{0.045}              \\
Ours               & \textbf{0.157435} & \eg{\textbf{0.049}}     & \textbf{0.138541} & \eg{\textbf{0.044}}     \\
\hline
\end{tabular}
\caption{The mean 3D mesh errors -- positional and normal, over all geometry reconstructions (lower is better).}
\label{tab:meshEvalAll}
  \vspace{-20px}
\end{table}
Finally, Table~\ref{tab:meshEvalAll} shows the mean geometric and normal error for reconstruction over twenty-four input images. Since, the positional distance metric does not measure \textit{smoothness or bumpiness} of the reconstructions, we report the normal distance error. The normal error computes the mean per-vertex unit-normal distance (on the unit sphere surface) between reconstruction and GT meshes.
\section{Mesh difference}
\label{sec:meshDiff}
The mean differences were computed per-vertex for each mesh. We implement a similar 3D mesh evaluation protocol as described in \cite{Pillai2019_3dfaw}. For computing the mesh difference, we first align the reconstructed mesh to GT meshes. Several feature points (sparse correspondence) are defined on both GT and the reconstructed face meshes, where vertices are minimally affected by facial muscles. With the corresponding sparse points on both meshes, we use a traditional least-square estimation introduced by~\cite{Umeyama1991LeastSquaresEO} to align the two meshes. After this alignment, we compute the distance from each vertex of a mesh to the other, and calculate the average of the distance measured by~\cite{Moller1997inter}.

%% file: 10.abblation-continued.tex
\section{Ablation Studies (continued)}
\begin{figure}
\includegraphics[width=\linewidth]{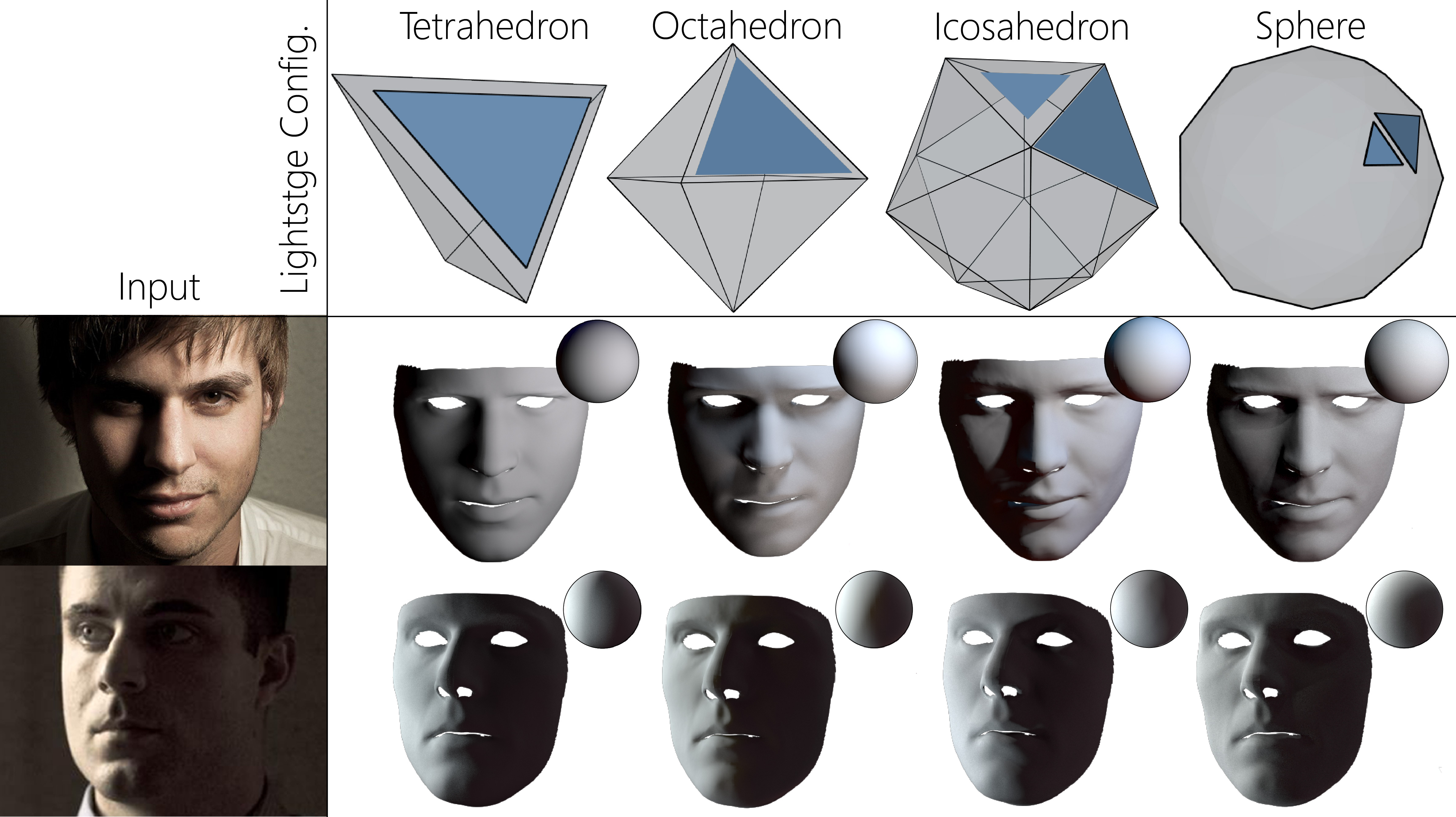}
\caption{Comparison results for different light-stage geometric configurations. For each subject, we show the estimated shadows and light direction (represented as sphere environment map).}
\label{fig:lightstage_ablation}
\vspace{-10px}
\end{figure}

\textbf{Light stage Geometries.} We compare different light stage configurations such as tetrahedron (four lights), octahedron (eight lights), icosahedron (twenty lights) and spherical (eighty lights) shape geometries, as shown on Figure~\ref{fig:lightstage_ablation}. For the topmost subject, the tetrahedon and octahedron light stage geometries fail to capture the bright area on the face (in the forehead area). The estimated light direction shows incoming light from the right (see the corresponding spherical environment map), while real light comes from the top-right side of the subject. The spherical light stage, provides a good approximation of the incoming light direction but the estimated shadows are not as accurate as the one produced by the icoshaedron.
The icoshaedron geometry, produces a high quality self-shadows approximation that is visually close to the input image, even the shadows on bottom of the subject's right eye are captured. For the second subject, all the light stage geometries provide a good approximation of the input self-shadows. We conclude that the icoshaedron provides the optimal setup for dataset of images shown in the paper. 
